\documentclass[5p]{elsarticle}
\usepackage{lineno, hyperref}
\usepackage{graphicx}
\usepackage{amsmath}
\usepackage{amssymb}
\usepackage{subfigure}
\usepackage{color}
\usepackage{epsfig}
\usepackage{textcomp}
\usepackage{url}
\usepackage[labelfont=bf,labelsep=quad,justification=justified]{caption}
\newcommand{\noit}[1]{\textup{#1}}
\usepackage[T1]{fontenc}
\usepackage{fix-cm}
\modulolinenumbers[5]

\journal{CVIU}
\def\etal{\emph{et al.}}

\begin{document}

\begin{frontmatter}

\title{Enhancing Energy Minimization Framework for\\ Scene Text Recognition with Top-Down Cues}
%\tnotetext[mytitlenote]{Fully documented templates are available in the elsarticle package on \href{http://www.ctan.org/tex-archive/macros/latex/contrib/elsarticle}{CTAN}.}

\author{Anand Mishra$^{1*}$}
\author{Karteek Alahari$^2$}
\author{C.~V.~Jawahar$^1$}
       \address{$^1$IIIT Hyderabad\fnref{cor1} \hspace{1cm} $^2$Inria\fnref{cor2}\\
       }
       \fntext[cor1]{Center for Visual Information Technology, IIIT Hyderabad, India.}
       \fntext[cor2]{LEAR team, Inria Grenoble Rhone-Alpes, Laboratoire Jean Kuntzmann, CNRS, Univ.\ Grenoble Alpes, France.\newline
\hspace*{0.22cm}*Corresponding author --\\
\hspace*{0.22cm}       Email:~\texttt{anand.mishra@research.iiit.ac.in}\\
\hspace*{0.22cm}       Phone: +91-40-6653 1000~~~Fax: +91-40-6653 1413}
       
%% Group authors per affiliation:
%\author{Anand Mishra}
%\address{Radarweg 29, Amsterdam}
%\fntext[myfootnote]{Since 1880.}
%
%%% or include affiliations in footnotes:
%\author[mymainaddress,mysecondaryaddress]{Elsevier Inc}
%\ead[url]{www.elsevier.com}
%
%\author[mysecondaryaddress]{Global Customer Service\corref{mycorrespondingauthor}}
%\cortext[mycorrespondingauthor]{Corresponding author}
%\ead{support@elsevier.com}
%
%\address[mymainaddress]{1600 John F Kennedy Boulevard, Philadelphia}
%\address[mysecondaryaddress]{360 Park Avenue South, New York}

\begin{abstract}
Recognizing scene text is a challenging problem, even more so than the
recognition of scanned documents. This problem has gained significant attention
from the computer vision community in recent years, and several methods based
on energy minimization frameworks and deep learning approaches have been
proposed. In this work, we focus on the energy minimization framework and
propose a model that exploits both bottom-up and top-down cues for recognizing
cropped words extracted from street images. The bottom-up cues are derived from
individual character detections from an image. We build a conditional random
field model on these detections to jointly model the strength of the detections
and the interactions between them. These interactions are top-down cues
obtained from a lexicon-based prior, i.e., language statistics. The optimal
word represented by the text image is obtained by minimizing the energy
function corresponding to the random field model. We evaluate our proposed
algorithm extensively on a number of cropped scene text benchmark datasets, namely
Street View Text, ICDAR 2003, 2011 and 2013 datasets, and IIIT 5K-word, and
show better performance than comparable methods. We perform a rigorous analysis
of all the steps in our approach and analyze the results. We also show that
state-of-the-art convolutional neural network features can be integrated in our
framework to further improve the recognition performance.
\end{abstract}

\end{frontmatter}

%\linenumbers
\section{Introduction}
\label{sec:intro}
\begin{figure}[!t]
\centering
\includegraphics[width=8cm,height=6cm]{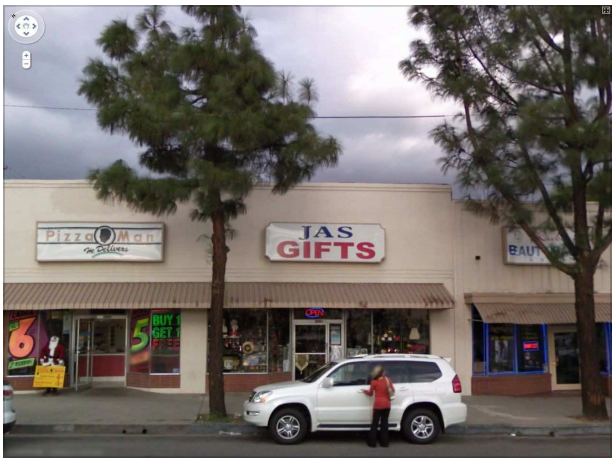}
\caption{A typical street scene image taken from Google Street View. It
contains very prominent sign boards with text on the building and its windows.
It also contains objects such as car, person, tree, and regions such as road,
sky. Many scene understanding methods recognize these objects and regions in
the image successfully, but overlook the text on the sign board, which contains
rich, useful information. The goal of this work is to address this gap in
understanding scenes.}
\label{fig:streetscene}
\end{figure}

The problem of understanding scenes semantically has been one of the
challenging goals in computer vision for many decades. It has gained
considerable attention over the past few years, in particular, in the context
of street scenes~\cite{Brostow08,Ladicky10,Geiger2012CVPR}. This problem has
manifested itself in various forms, namely, object
detection~\cite{Desai09,Felzen10}, object recognition and
segmentation~\cite{Levin09,Shotton09}. There have also been significant
attempts at addressing all these tasks
jointly~\cite{Ladicky10,Gould09b,YaoFU12}. Although these approaches interpret
most of the scene successfully, regions containing text are overlooked. As an
example, consider an image of a typical street scene taken from Google Street
View in Fig.~\ref{fig:streetscene}. One of the first things we notice in this
scene is the sign board and the text it contains. However, popular recognition
methods ignore the text, and identify other objects such as car, person, tree,
and regions such as road, sky. The importance of text in images is also
highlighted in the experimental study conducted by Judd~\textit{et
al.}~\cite{Judd09}. They found that viewers fixate on text when shown images
containing text and other objects. This is further evidence that text
recognition forms a useful component in understanding scenes.

%------------------------------------------------------------------------
\begin{figure*}[!t]
\centering
\begin{tabular}{cccccc}
\includegraphics[width=2.3cm,height=1.2cm]{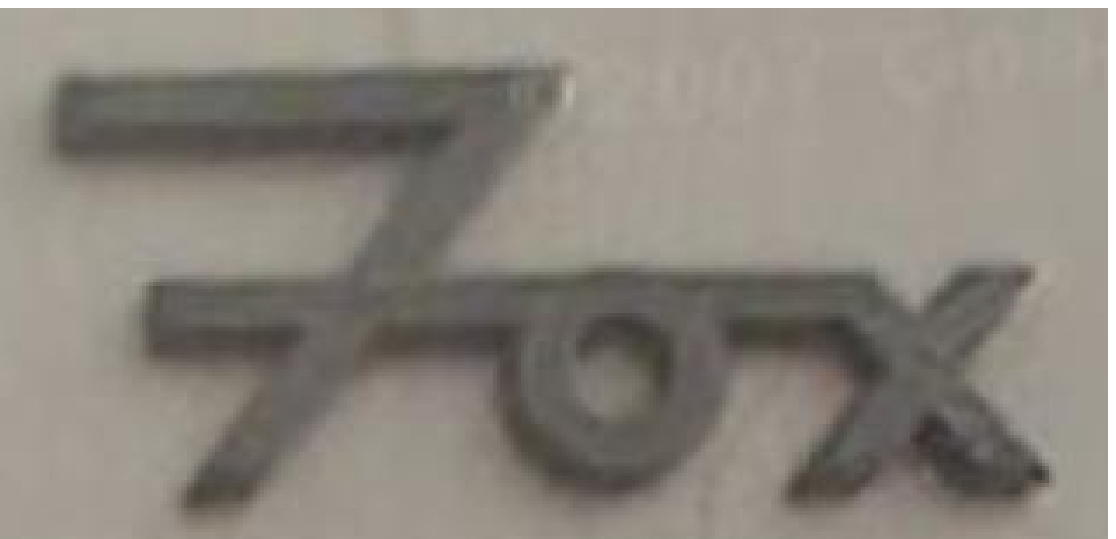} &
\includegraphics[width=2.3cm,height=1.2cm]{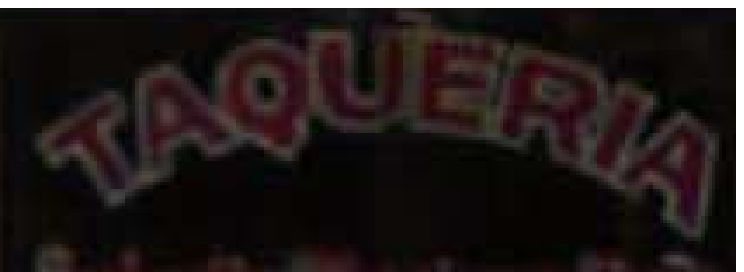} &
\includegraphics[width=2.3cm,height=1.2cm]{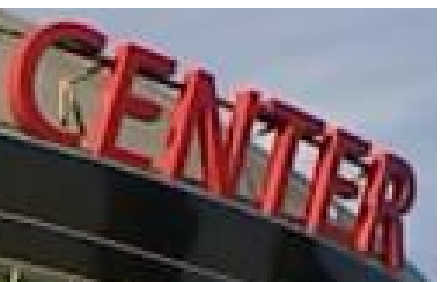} &
\includegraphics[width=2.3cm,height=1.2cm]{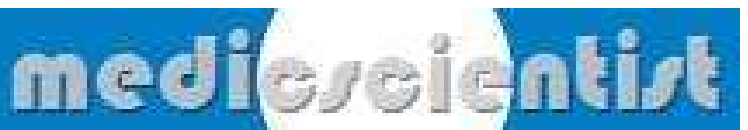}&
\includegraphics[width=2.3cm,height=1.2cm]{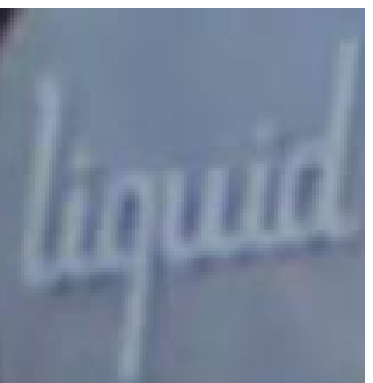}&
\includegraphics[width=2.3cm,height=1.2cm]{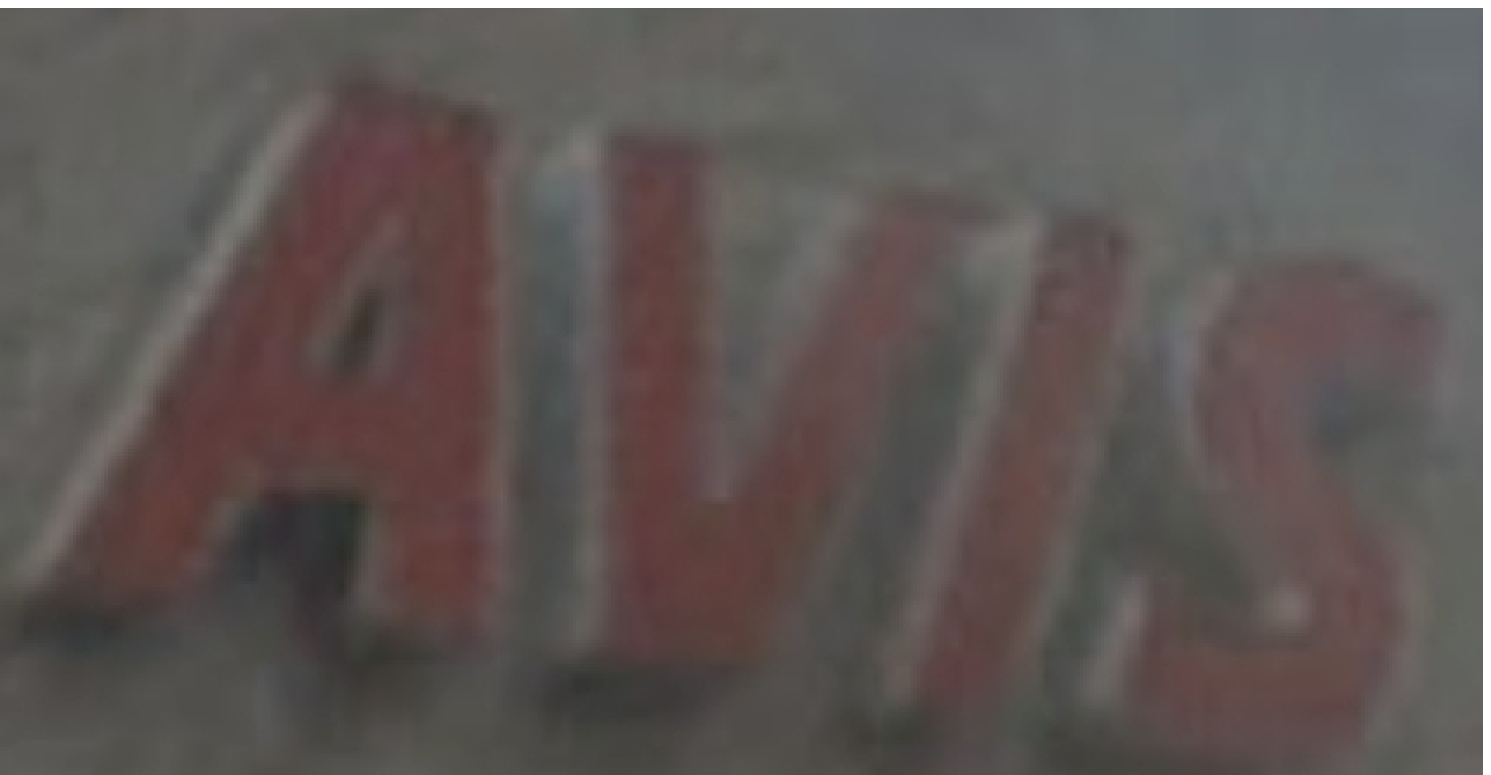}\\

\includegraphics[width=2.3cm,height=1.2cm]{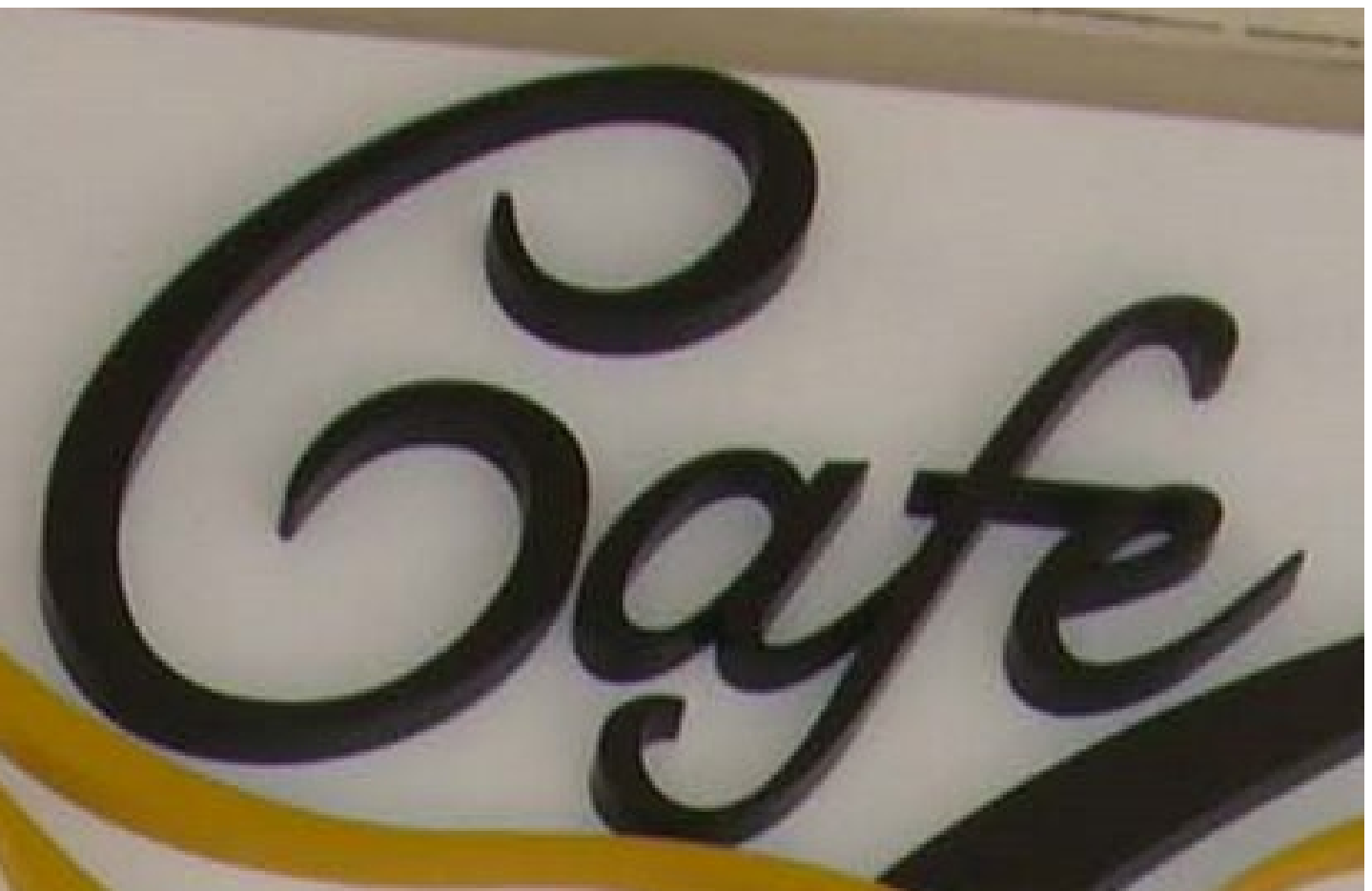}&
\includegraphics[width=2.3cm,height=1.2cm]{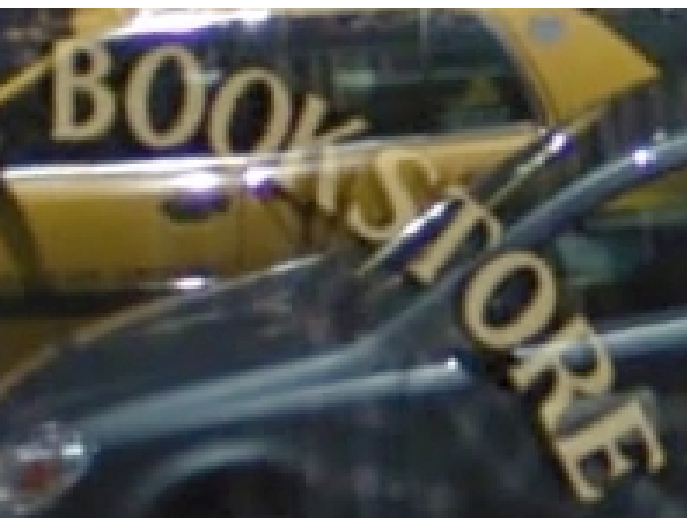}&
\includegraphics[width=2.3cm,height=1.2cm]{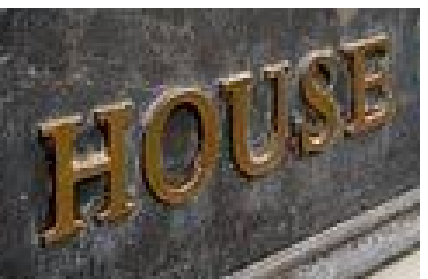}&
\includegraphics[width=2.3cm,height=1.2cm]{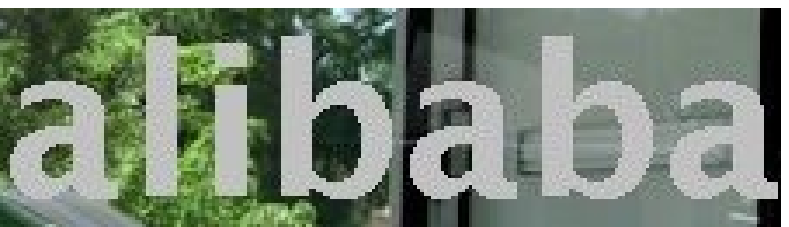}&
\includegraphics[width=2.3cm,height=1.2cm]{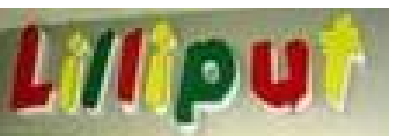}&
\includegraphics[width=2.3cm,height=1.2cm]{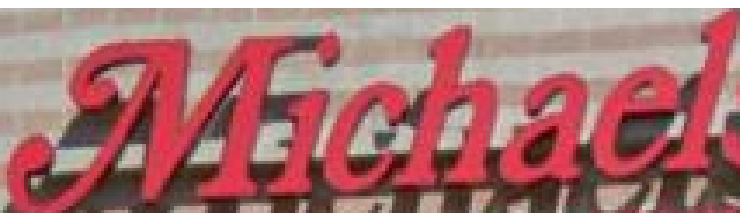}\\
\end{tabular}
\caption{Challenges in scene text recognition. A few sample images from the SVT
and IIIT 5K-word datasets are shown to highlight the variation in view point,
orientation, non-uniform background, non-standard font styles and also issues
such as occlusion, noise, and inconsistent lighting. Standard OCRs perform
poorly on these datasets (as seen in Table~\ref{tab:ourdataset}
and~\cite{WangB11,NeumannM10}).}
\label{fig:challenges2}
\end{figure*}

In addition to being an important component of scene understanding, scene text
recognition has many potential applications, such as image retrieval, auto
navigation, scene text to speech systems, developing apps for visually impaired
people~\cite{Mishra13,text2speech}. Our method for solving this task is
inspired by the many advancements made in the object detection and recognition
problems~\cite{Desai09,Felzen10,Shotton09,DalalT05}. We present a framework for
recognizing text that exploits bottom-up and top-down cues. The bottom-up cues
are derived from individual character detections from an image. Naturally,
these windows contain true as well as false positive detections of characters.
We build a conditional random field (CRF) model~\cite{Lafferty01} on these
detections to determine not only the true positive detections, but also the
word they represent jointly. We impose top-down cues obtained from a
lexicon-based prior, i.e., language statistics, on the model. In addition to
disambiguating between characters, this prior also helps us in recognizing
words.

The first contribution of this work is a joint framework with seamless
integration of multiple cues---individual character detections and their
spatial arrangements, pairwise lexicon priors, and higher-order priors---into a CRF
framework which can be optimized effectively. The proposed method performs
significantly better than other related energy minimization based methods for
scene text recognition. Our second contribution is devising a cropped word recognition
framework which is applicable not only to closed vocabulary text recognition
(where a small lexicon containing the ground truth word is provided with each
image), but also to a more general setting of the problem, i.e., open
vocabulary scene text recognition (where the ground truth word may or may not
belong to a generic large lexicon or the English dictionary). The third
contribution is comprehensive experimental evaluation, in contrast to many
recent works, which either consider a subset of benchmark datasets or are
limited to the closed vocabulary setting. We evaluate on a number of cropped word datasets (ICDAR 2003, 2011 and 2013~\cite{ICDAR}, SVT~\cite{SVT}, and IIIT
5K-word~\cite{MishraBMVC12}) and show results in closed and open vocabulary
settings. Additionally, we analyzed the effectiveness of individual components
of the framework, the influence of parameter settings, and the use of
convolutional neural network (CNN) based features~\cite{AZ14}.

The remainder of the paper is organized as follows. In
Section~\ref{sec:relWork} we discuss related work. Section~\ref{sec:recMod}
describes our scene text recognition model and its components. We then present
the evaluation protocols and the datasets used in experimental analysis in
Section~\ref{datasets}. Comparison with related approaches is shown in
Section~\ref{sec:expts}, along with implementation details. We then make
concluding remarks in Section~\ref{sec:conclusion}.

\section{Related Work}
\label{sec:relWork}
The task of understanding scene text has gained a huge interest for more than a
decade~\cite{WangB11,NeumannM10,EpshteinW10,WeinmanLH09,weinman2013toward,pushmeetLexi,JoseBMVC13,fieldICDAR13,CamposBV09,ChenOB02,neumann2012real,photoOCR,AZ14,stroklets}.
It is closely related to the problem of Optical Character Recognition (OCR),
which has a long history in the computer vision and pattern recognition
communities~\cite{TwentyYears}. However, the success of OCR systems is largely
restricted to text from scanned documents. Scene text exhibits a large
variability in appearance, as shown in Fig.~\ref{fig:challenges2}, and can
prove to be challenging even for the state-of-the-art OCR methods (see
Table~\ref{tab:ourdataset} and~\cite{WangB11,NeumannM10}). The problems in this
context are: (1) text localization, (2) cropped word recognition, and (3)
isolated character recognition. They have been tackled either
individually~\cite{EpshteinW10,CamposBV09,ChenY04}, or
jointly~\cite{WangB11,AZ14,weinman2013toward,neumann2012real}. This paper
focuses on addressing the cropped word recognition problem. In other words,
given an image region (e.g., in the form of a bounding box) containing text,
the task is to recognize this content. The core components of a typical cropped
word recognition framework are: localize the characters, recognize them, and use
statistical language models to compose the characters into words. Our framework
builds on these components, but differs from previous work in several ways. In
the following, we review the prior art and highlight these differences. The
reader is encouraged to refer to~\cite{PAMIsurvey} for a more comprehensive
survey of scene text recognition methods.

\begin{figure}[!t]
\centering
\subfigure{
\label{fig:subfig3}
\includegraphics[height=1.8cm,width=3.5cm]{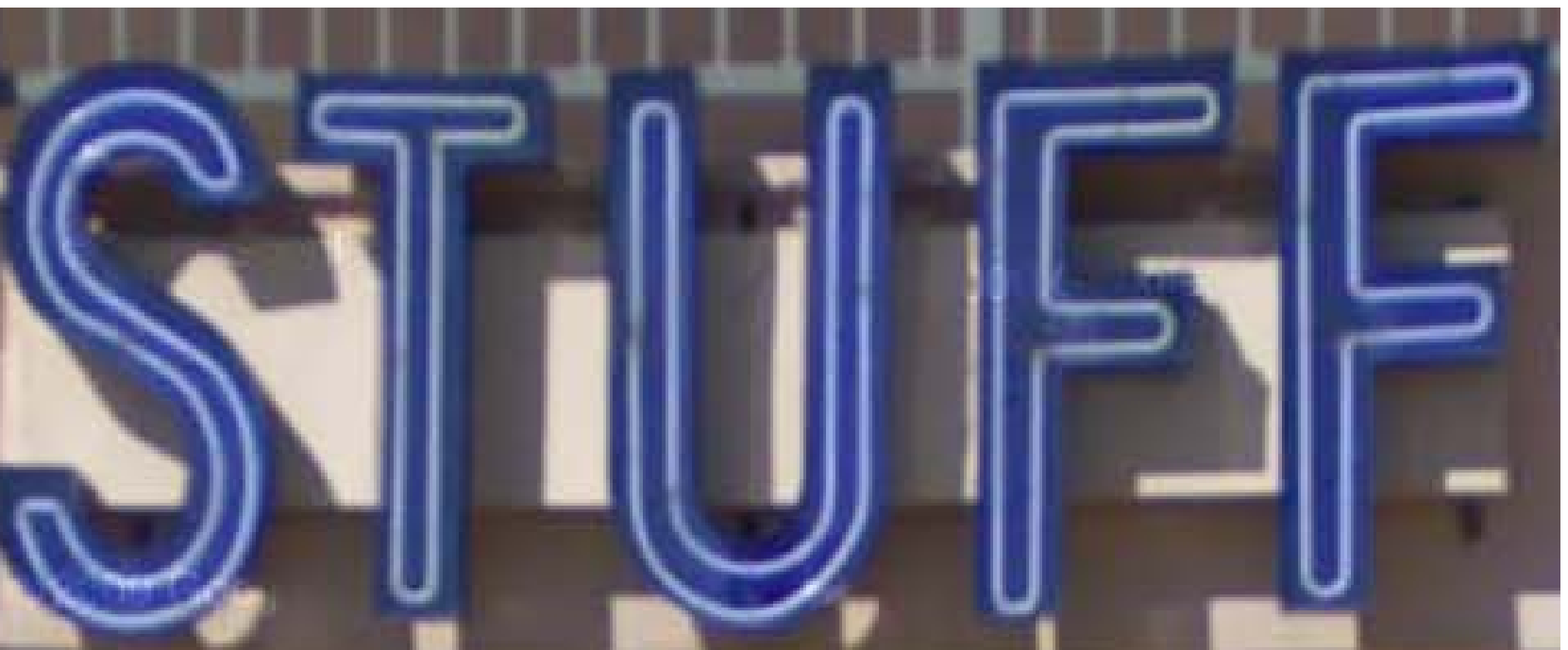}
%\hspace{0.1cm}
\includegraphics[height=1.8cm,width=3.5cm]{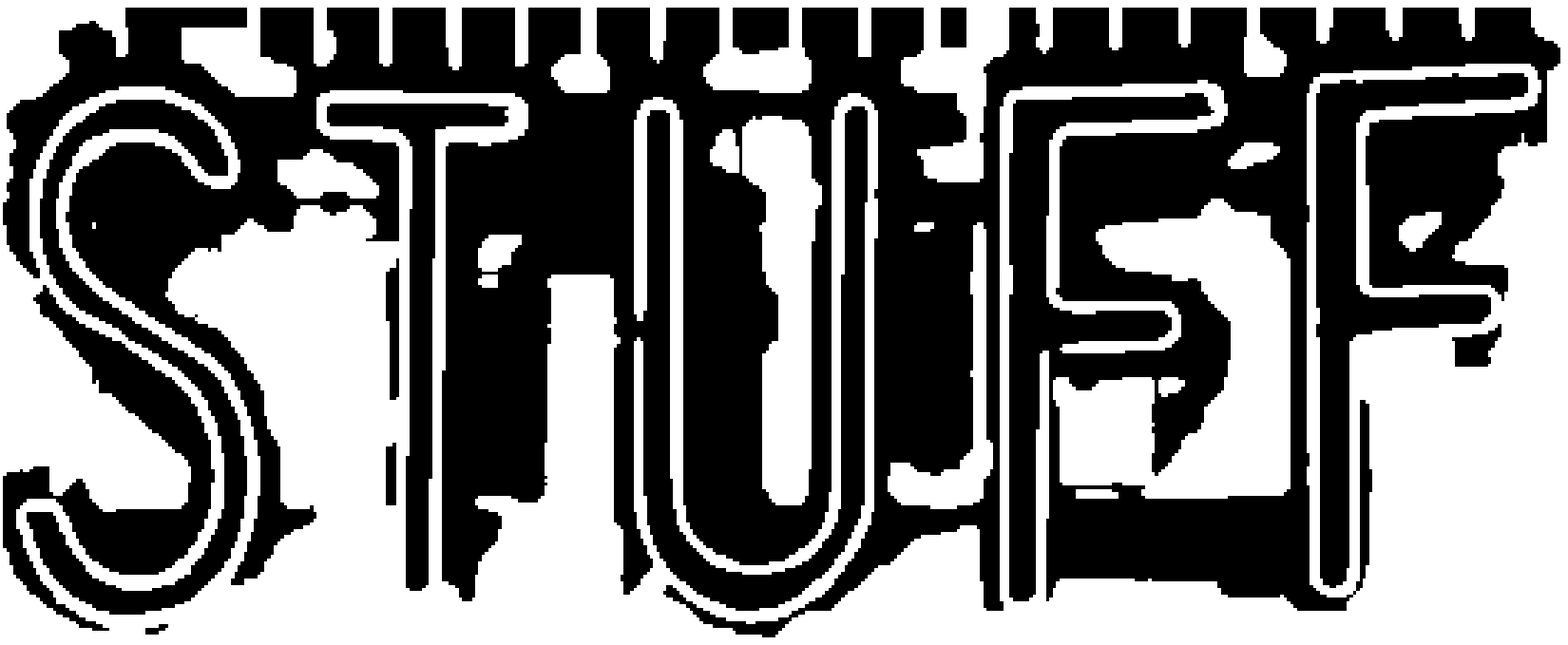}
}
\hspace{0.8cm}
\subfigure{
\label{fig:subfig4}
\includegraphics[height=1.8cm,width=3.5cm]{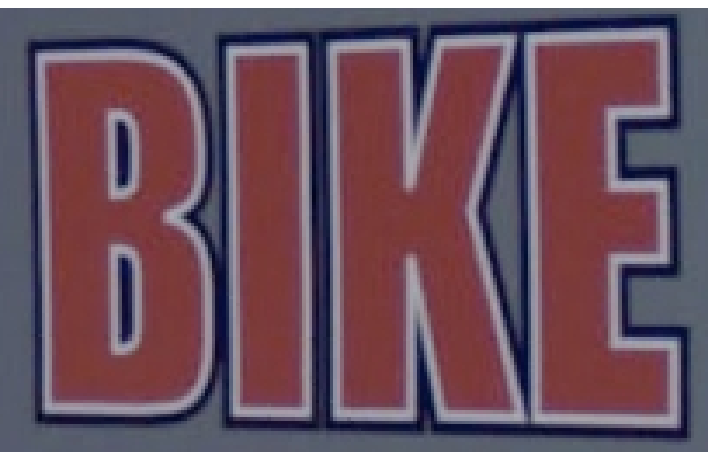}
%\hspace{0.1cm}
\includegraphics[height=1.8cm,width=3.5cm]{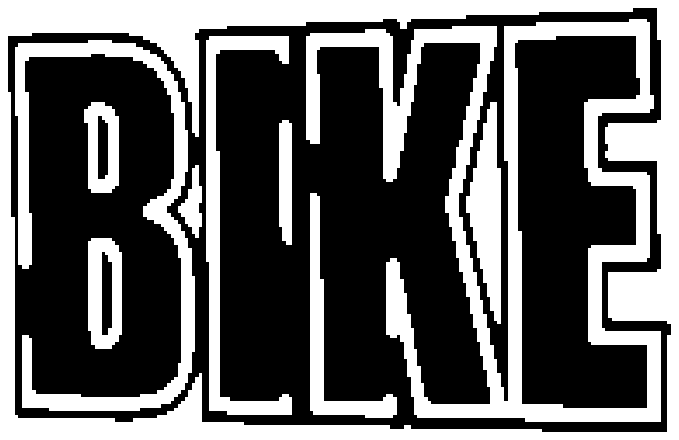}
}
\caption{Binarization results obtained with one of the state-of-the-art
methods~\cite{MishraJ11} are shown for two sample images. We observed similar
poor performance on most of the images in scene text datasets, and hence do not
use binarization in our framework.}
\label{fig:challenges1}
\end{figure}

A popular technique for localizing characters in an OCR system is to
binarize the image and determine the potential character locations based on
connected components~\cite{OCRlexi}. Such techniques have also been adapted for
scene text recognition~\cite{NeumannM10}, although with limited success. This
is mainly because obtaining a clean binary output for scene text images is
often challenging; see Fig.~\ref{fig:challenges1} for examples. An
alternative approach is proposed in~\cite{shivakumarICDAR11} using gradient
information to find potential character locations. More recently,
Yao~\textit{et al.}~\cite{stroklets} proposed a mid-level feature based
technique to localize characters in scene text. We follow an alternative strategy
and cast the character localization problem as an object detection task, where
characters are the {\it objects}.  We then define an energy function on all the
potential characters.% However, we believe more robust character localization technique can further enhance our word recognition performance.

One of the earliest works on large-scale natural scene character recognition
was presented in~\cite{CamposBV09}. This work develops a multiple kernel
learning approach using a set of shape-based features. Recent
work~\cite{WangB11,MishraCVPR12} has improved over this with histogram of
gradient features~\cite{DalalT05}. We perform an extensive analysis on
features, classifiers, and propose methods to improve character recognition
further, for example, by augmenting the training set. In addition to this, we
show that the state-of-the-art CNN features~\cite{AZ14} can be successfully
integrated with our word recognition framework to further boost its
performance.

A study on human reading psychology shows that our reading improves
significantly with prior knowledge of the language~\cite{bookreading}.
Motivated by such studies, OCR systems have used, often in post-processing
steps~\cite{OCRlexi,tongOCRLexi}, statistical language models like $n$-grams to
improve their performance. Bigrams or trigrams have also been used in the
context of scene text recognition as a post-processing step,
e.g.,~\cite{STRlexi}. A few other
works~\cite{thillou2005,elagouni2011,elagouni2012combining} integrate character
recognition and linguistic knowledge to deal with recognition errors. For
example,~\cite{thillou2005} computes $n$-gram probabilities from more than 100
million characters and uses a Viterbi algorithm to find the correct word.  The
method in~\cite{elagouni2012combining}, developed in the same year as our CVPR
2012 work~\cite{MishraCVPR12}, builds a graph on potential character locations
and uses $n$-gram scores to constrain the inference algorithm to predict the
word. In contrast, our approach uses a novel location-specific prior
(cf.~(\ref{eq:lexicon2})).
%In contrast, we integrate the language models directly
%into our framework, as part of the energy function.

The word recognition problem has been looked at in two contexts---
with~\cite{WangB11,JoseBMVC13,MishraCVPR12,ngICPR12,GoelICDAR13} and
without~\cite{WeinmanLH09,MishraBMVC12,weinmanLH08} the use of an
image-specific lexicon. In the case of image-specific lexicon-driven word
recognition, also known as the closed vocabulary setting, a list of words is
available for every scene text image. The task of recognizing the word now
reduces to that of finding the best match from this list. This is relevant in
many applications, e.g., recognizing text in a grocery store, where a list of
grocery items can serve as a lexicon. Wang~\etal~\cite{ngICPR12} adapted a
multi-layer neural network for this scenario. In~\cite{WangB11}, each word in
the lexicon is matched to the detected set of character windows, and the one
with the highest score is reported as the predicted word. In one of our
previous works~\cite{GoelICDAR13}, we compared features computed on the entire
scene text image and those generated from synthetic font renderings of lexicon
words with a novel weighted dynamic time warping (wDTW) approach to recognize
words. In~\cite{JoseBMVC13} Rodriguez-Serrano and Perronnin proposed to embed
word labels and word images into a common Euclidean space, wherein the text
recognition task is posed as a retrieval problem to find the closest word label
for a given word image. While all these approaches are interesting, their
success is largely restricted to the closed vocabulary setting and cannot be
easily extended to the more general cases, for instance, when image-specific
lexicon is unavailable. Weinman~\etal~\cite{WeinmanLH09} proposed a method to
address this issue, although with a strong assumption of known character
boundaries, which are not trivial to obtain with high precision on the datasets
we use. The work in~\cite{weinmanLH08} generalizes their previous approach by
relaxing the character-boundary requirement. It is, however, evaluated only on
``roughly fronto-parallel'' images of signs, which are less challenging than
the scene text images used in our work. 

Our work belongs to the class of word recognition methods
which build on individual character localization, similar to methods such
as~\cite{NeumannM10,HoweFM09}. In this framework, the potential characters are
localized, then a graph is constructed from these locations, and then the
problem of recognizing the word is formulated as finding an optimal path in
this graph~\cite{NeumannM13} or inferring from an ensemble of
HMMs~\cite{HoweFM09}. Our approach shows a seamless integration of higher order
language priors into the graph (in the form of a CRF model), and uses
more effective modern computer vision features, thus making it clearly
different from previous works.

Since the publication of our original work in CVPR 2012~\cite{MishraCVPR12} and
BMVC 2012~\cite{MishraBMVC12} papers, several approaches for scene text
understanding (e.g., text
localization~\cite{Milyaev13,neumann2012real,Jaderberg14a,Huang14}, word
recognition~\cite{AZ14,weinman2013toward,photoOCR,stroklets,shiCVPR13,Jaderberg14a}
and text-to-image retrieval~\cite{Mishra13,Jaderberg14a,AlmazanGFV14,Roy14})
have been proposed. Notably, there has been an increasing interest in exploring
deep convolutional network based methods for scene text tasks
(see~\cite{AZ14,photoOCR,ngICPR12,Jaderberg14a,Huang14} for example). These
approaches are very effective in general, but the deep convolutional network,
which is at the core of these approaches, lacks the capability to elegantly
handle structured output data. To understand this with the help of an example,
let us consider the problem of estimating human
pose~\cite{toshev2013deeppose,Tompson14}, where the task is to predict the
locations of human body joints such as head, shoulders, elbows and wrists.
These locations are constrained by human body kinematics and in essence form a
structured output. To deal with such structured output data, state-of-the-art
deep learning algorithms include an additional regression
step~\cite{toshev2013deeppose} or a graphical model~\cite{Tompson14}, thus
showing that these techniques are complementary to the deep learning
philosophy. Similar to human pose, text is structured output
data~\cite{Jaderberg14b}. To better handle this structured data, we develop our
energy minimization framework~\cite{MishraBMVC12,MishraCVPR12} with the
motivation of building a complementary approach, which can further benefit
methods built on the deep learning paradigm. Indeed, we see that combining the
two frameworks further improves text recognition results
(Section~\ref{sec:expts}).
%\textcolor{green} {This extended version includes an improved character classifier, analyzes the influence of the lexicon size on the text recognition performance, shows results on various public benchmarks in both open and closed vocabulary setting and compares it with many recent works.}
\section{The Recognition Model} 
\label{sec:recMod}
We propose a conditional random field (CRF) model for recognizing words. The
CRF is defined over a set of $N$ random variables $x = \{x_{i} |i \in
\mathcal{V}\}$, where $\mathcal{V} = \{1, 2, \ldots, N\}$. Each random variable
$x_i$ denotes a potential character in the word, and can take a label from the
label set $\mathcal{L} = \{l_{1}, l_{2}, \ldots, l_{k}\} \cup \epsilon$, which
is the set of English characters, digits and a null label $\epsilon$ to discard
false character detections. The most likely word represented by the set of
characters $x$ is found by minimizing the energy function, $E:\mathcal{L}^{n}
\rightarrow \mathbb{R}$,
corresponding to the random field. The energy function $E$ can be 
written as sum of potential functions:
\begin{equation}
E(x)= \sum_{c \in \cal{C}}{\psi_{c} (x_{c})},
\label{eqn:energy}
\end{equation}
where $\cal{C} \subset \mathcal{P}(\mathcal{V})$, with
$\mathcal{P}(\mathcal{V})$ denoting the powerset of $\mathcal{V}$. Each $x_{c}$
defines a set of random variables included in subset $c$, referred to as a
clique. The function $\psi_c$ defines a constraint (potential) on the
corresponding clique $c$. We use unary, pairwise and higher order potentials in
this work, and define them in Section~\ref{sec:GC}. The set of potential
characters is obtained by the character detection step discussed in
Section~\ref{sec:charDet}. The neighbourhood relations among characters,
modelled as pairwise and higher order potentials, are based on the spatial
arrangement of characters in the word image.

In the following we show an example energy function composed of unary, pairwise
and higher order (of clique size three) terms on a sample word with four
characters. For a word to be recognized as ``OPEN'' the following energy
function should be the minimum.
\begin{equation}
\begin{aligned}
\psi(O,P,E,N)& = \psi_{1}(O)+\psi_{1}(P)+\psi_{1}(E)+\psi_{1}(N)\nonumber\\
&\quad + \psi_{2}(O,P) + \psi_{2}(P,E) + \psi_{2}(E,N)\\
&\quad + \psi_3(O,P,E) + \psi_3(P,E,N).
\end{aligned}
\end{equation}
The third order terms $\psi_3(O,P,E)$ and $\psi_3(P,E,N)$ are decomposed as
follows.
\begin{equation}
\begin{aligned}
\psi_3(O,P,E) & = \psi^a_1(OPE) + \psi^a_2(OPE,O) \nonumber \\
              &\quad + \psi^a_2(OPE,P) + \psi^a_2(OPE,E).
\end{aligned}
\end{equation}
\begin{equation}
\begin{aligned}
\psi_3(P,E,N)& = \psi^a_1(PEN) + \psi^a_2(PEN,P)\nonumber\\
             &\quad + \psi^a_2(PEN,E) + \psi^a_2(PEN,N).
\end{aligned}
\end{equation}
\subsection{Character Detection}
\label{sec:charDet}
The first step in our approach is to detect potential locations of characters
in a word image. In this work we use a sliding window based approach for
detecting characters, but other methods, e.g.,~\cite{stroklets}, can also be
used instead.

\begin{figure}[!t]
%\centring
\subfigure{
\label{fig:subfig1}
\includegraphics[height=1.5cm,width=3cm]{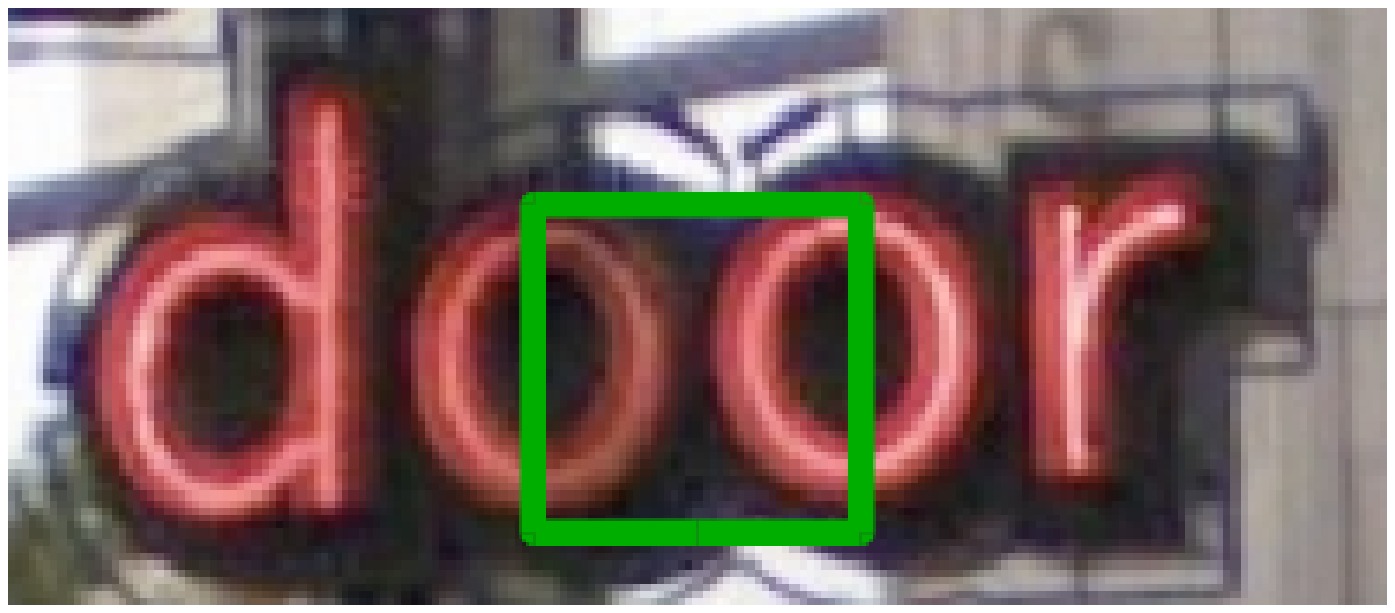}
%\hspace{0.1cm}
\includegraphics[height=1.0cm,width=0.6cm]{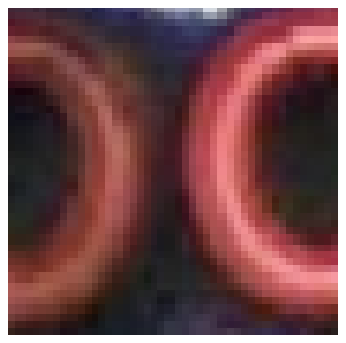}
}
\hspace{0.4cm}
\subfigure{
\label{fig:subfig2}
\includegraphics[height=1.5cm,width=3cm]{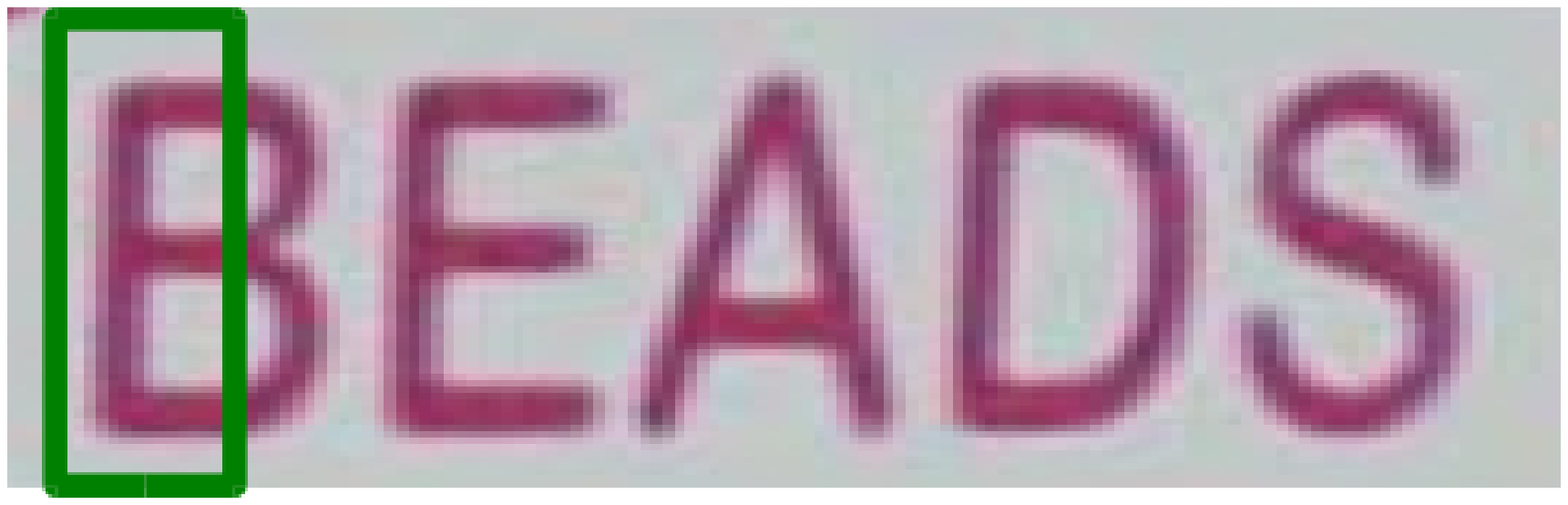}
%\hspace{0.1cm}
\includegraphics[height=1.5cm,width=0.5cm]{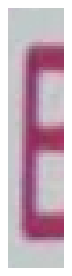}
}
\caption{Typical challenges in character detection. (a) Inter-character
confusion: A window containing parts of the two $o$'s is falsely detected as
$x$. (b) Intra-character confusion: A window containing a part of the character
B is recognized as E.}
\label{fig:challenges3}
\end{figure}

\paragraph{Sliding window detection}
This technique has been very successful for tasks such as, face~\cite{Viola01}
and pedestrian~\cite{DalalT05} detection, and also for recognizing handwritten
words using HMM based methods~\cite{hmm1}. Although character detection in
scene images is similar to such problems, it has its unique challenges.
Firstly, there is the issue of dealing with many categories ($63$ in all)
jointly. Secondly, there is a large amount of inter-character and
intra-character confusion, as illustrated in Fig.~\ref{fig:challenges3}. When a
window contains parts of two characters next to each other, it may have a very
similar appearance to another character. In Fig.~\ref{fig:challenges3}(a), the
window containing parts of the characters `$o$' can be confused with `$x$'.
Furthermore, a part of one character can have the same appearance as that of
another. In Fig.~\ref{fig:challenges3}(b), a part of the character `B' can be
confused with `E'. We build a robust character classifier and adopt an
additional pruning stage to overcome these issues.

The problem of classifying natural scene characters typically suffers from the
lack of training data, e.g., \cite{CamposBV09} uses only 15 samples per class.
It is not trivial to model the large variations in characters using only a few
examples. To address this, we add more examples to the training set by applying
small affine transformations~\cite{simardNIPS91,MozerNIPS97} to the original
character images. We further enrich the training set by adding many
non-character negative examples, i.e., from the background. With this strategy,
we achieve a significant boost in character classification accuracy (see
Table~\ref{tab:charClassification}).

We consider windows at multiple scales and spatial locations. The location of
the $i$th window, $d_i$, is given by its center and size. The set $\mathcal{K}
= \{c_1, c_2, \ldots, c_k\}$, denotes label set. Note that $k=63$ for the set
of English characters, digits and a background class (null label) in our work.
Let $\phi_{i}$ denote the features extracted from a window location $d_i$.
Given the window $d_i$, we compute the likelihood, $p(c_j|\phi_i)$, of it
taking a label $c_j$ for all the classes in $\mathcal{K}$. In our
implementation, we used explicit feature representation~\cite{VedaldiPAMI12} of
histogram of gradient (HOG) features~\cite{DalalT05} for $\phi_i$, and the
likelihoods $p$ are (normalized) scores from a one vs rest multi-class support
vector machine (SVM). Implementation details of the training procedure are
provided in Section~\ref{subsec:charclassif}.

This basic sliding window detection approach produces many potential character
windows, but not all of them are useful for recognizing words. We discard some
of the weak detection windows using the following pruning method.
\begin{figure}[!t]
\centering
\subfigure[]
{
 \includegraphics[scale=0.2]{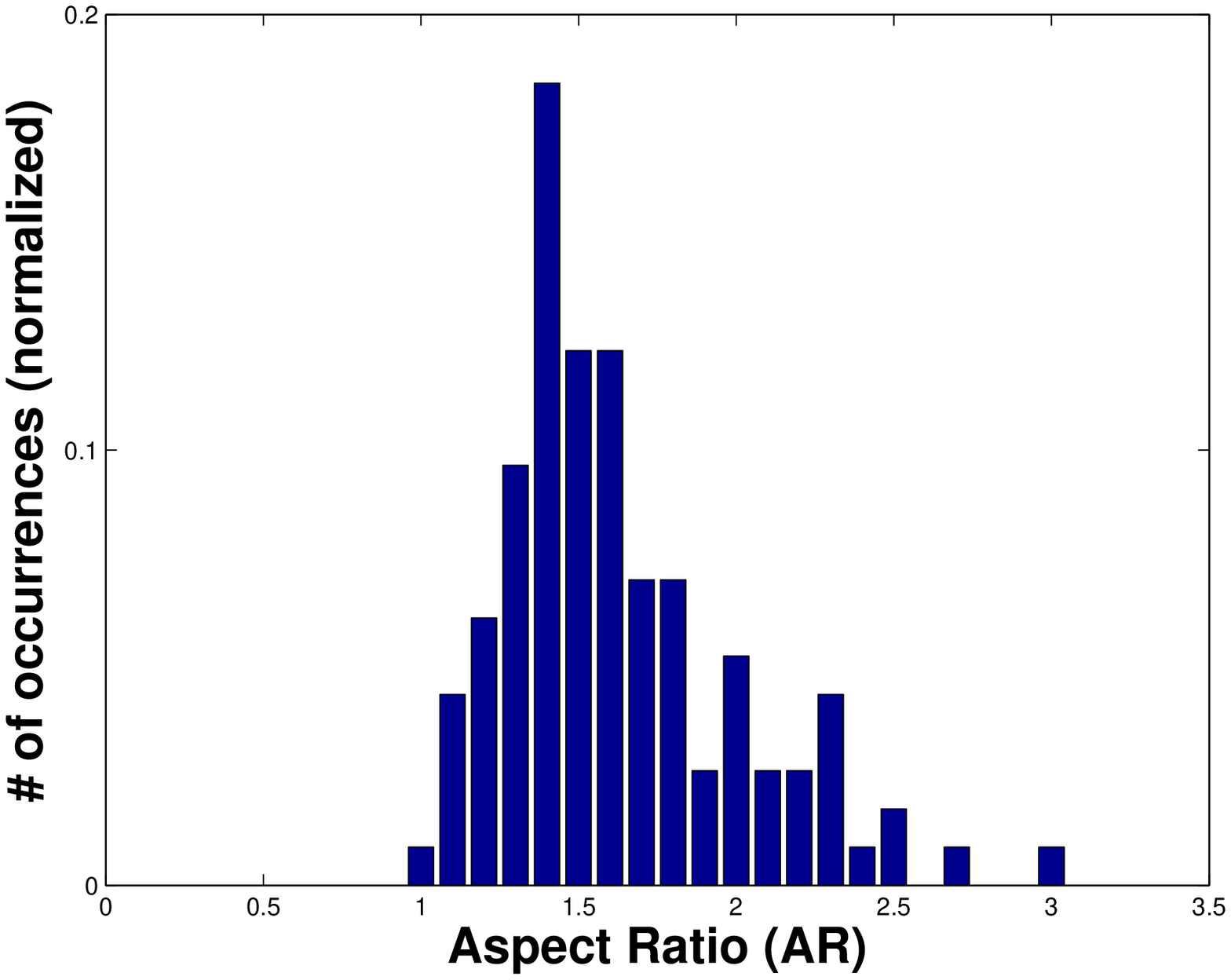}
}
\subfigure[]
{
\includegraphics[scale=0.2]{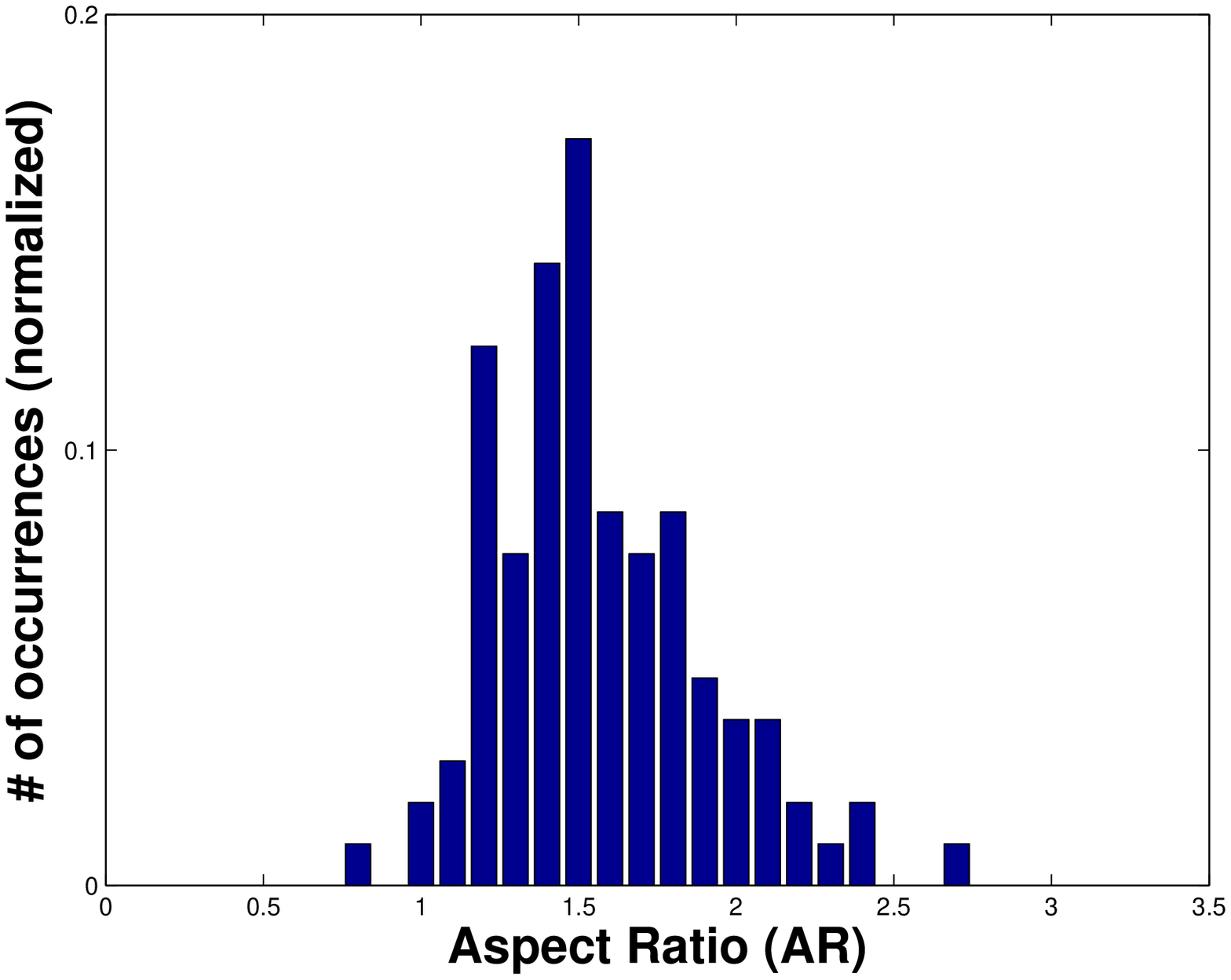}
} 
\subfigure[]
{
 \includegraphics[scale=0.2]{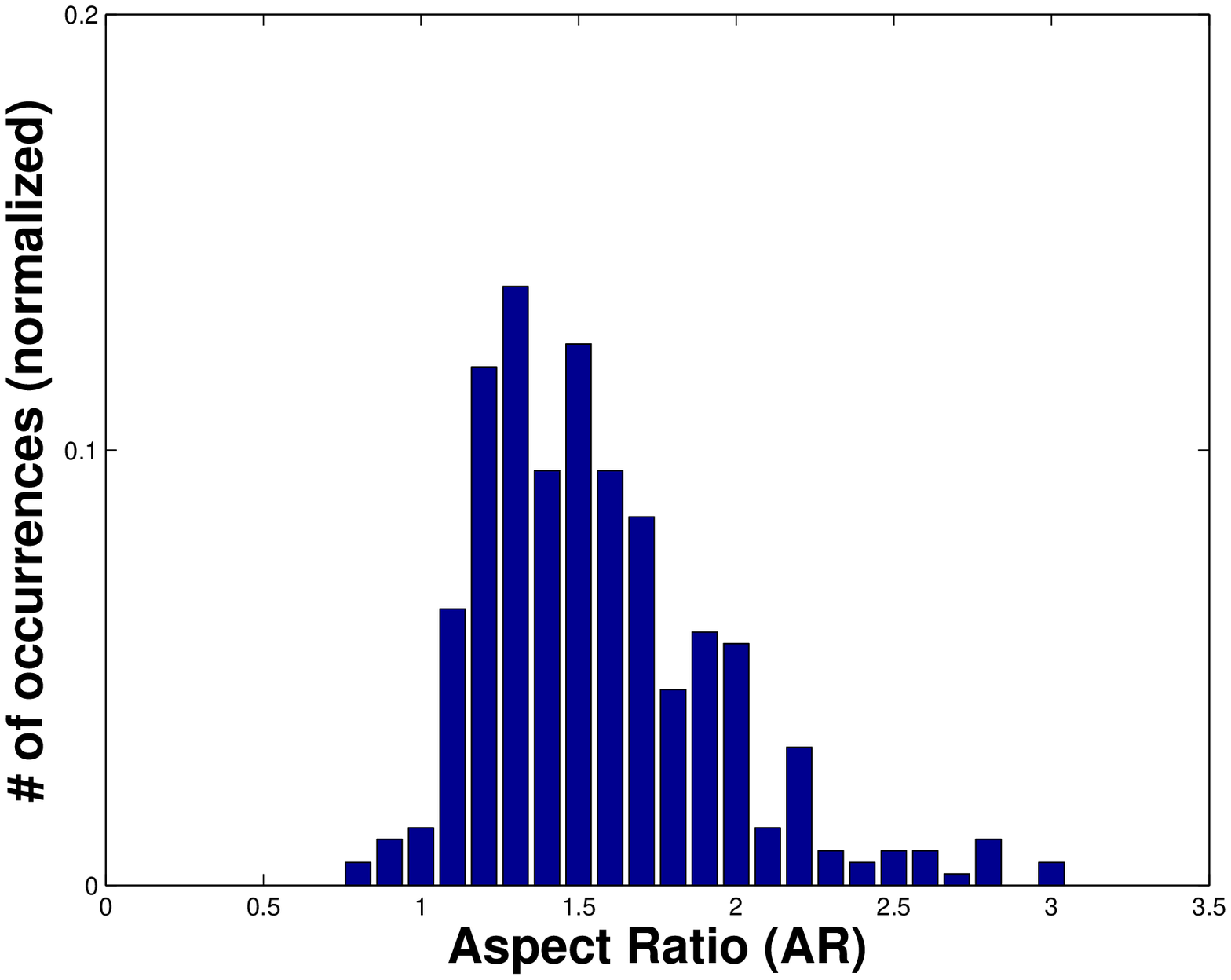}
}
\subfigure[]
{
\includegraphics[scale=0.2]{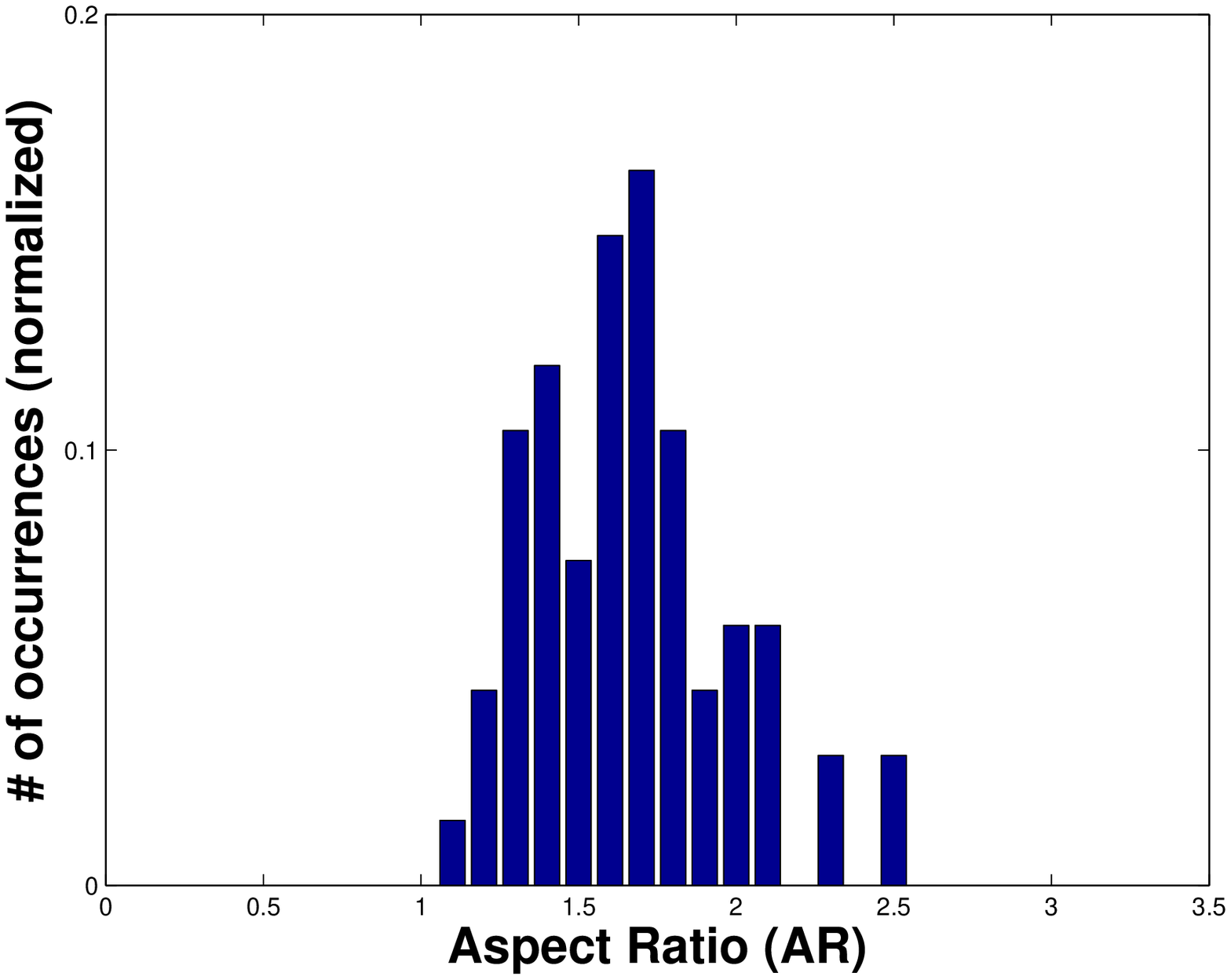}
} 
\caption{Distribution of aspect ratios of few digits and characters: (a) 0 (b)
2 (c) B (d) Y. The aspect ratios are computed on character from the IIIT-5K
word training set.}
\label{fig:ar}
\end{figure}

\begin{figure*}[!t]
\centering
 \includegraphics[scale=0.5]{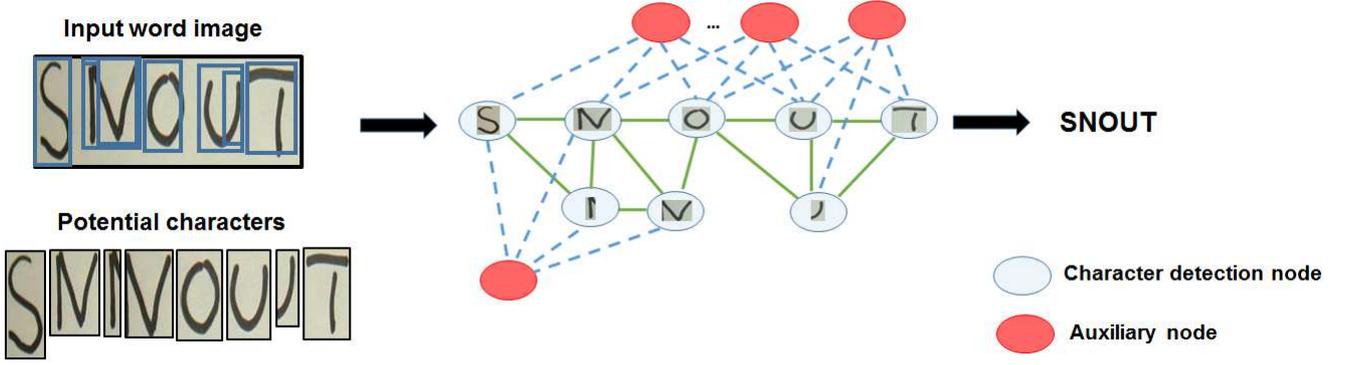}
\caption{The proposed model illustrated as a graph. Given a word image (shown
on the left), we evaluate character detectors and obtain potential character
windows, which are then represented in a graph. These nodes are connected with
edges based on their spatial positioning. Each node can take a label from the
label set containing English characters, digits, and a null label (to suppress
false detections). To integrate language models, i.e., $n$-grams, into the
graph, we add auxiliary nodes (shown in red), which constrain several character
windows together (sets of $4$ characters in this example). Auxiliary nodes take
labels from a label set containing all valid English $n$-grams and an
additional label to enforce high cost for an invalid $n$-gram.}
\label{fig:pgm}
\end{figure*}

\paragraph{Pruning windows}
For every potential character window, we compute a score based on: (i) SVM
classifier confidence, and (ii) a measure of the aspect ratio of the character
detected and the aspect ratio learnt for that character from training data. The
intuition behind this score is that, a strong character window candidate should
have a high classifier confidence score, and must fall within some range of the
sizes observed in the training data. In order to define the aspect ratio
measure, we observed the distribution of aspect ratios of characters from the
IIIT-5K word training set. A few examples of these distributions are shown in
Fig.~\ref{fig:ar}. Since they follow a Gaussian distribution, we chose this
score accordingly. For a window $d_i$ with an aspect ratio $a_i$, let $c_j$
denote the character with the best classifier confidence value given by
$S_{ij}$. The mean aspect ratio for the character $c_j$ computed from training
data is denoted by $\mu_{a_j}$. We define a goodness score (GS) for the window
$d_i$ as:
\begin{equation}
\text{GS}(d_{i}) = S_{ij} \exp \left (-\frac{(\mu_{a_{j}} - a_{i})^{2}}{2\sigma_{a_{j}}^2} \right ),
\label{eq:gs}
\end{equation}
where $\sigma_{a_{j}}$ is the variance of the aspect ratio for character
$c_{j}$ in the training data. A low goodness score indicates a weak detection,
which is then removed from the set of candidate character windows.

We then apply character-specific non-maximum suppression (NMS), similar to
other sliding window detection methods~\cite{Felzen10}, to address the issue of
multiple overlapping detections for each instance of a character. In other
words, for every character class, we select detections which have a high
confidence score, and do not overlap significantly with any of the other
stronger detections of the same character class. We perform NMS after aspect
ratio pruning to avoid wide windows with many characters suppressing weaker
single character windows they overlap with. The pruning and NMS steps are
performed conservatively, to discard only the obvious false detections. The
remaining false positives are modelled in an energy minimization framework with
language priors and other cues, as discussed below.
%~\textcolor{red}{We empirically 
%verify pruning step is indeed useful to obtain a good accuracy.}

\subsection{Graph Construction and Energy Formulation}
\label{sec:GC}
%\begin{figure}[!t]
%\centering
%\includegraphics[width=5cm,height=4cm]{figures/example.eps}
%\caption{\it Graph construction. We first find a set of potential character
%windows, shown at the top (only a few are shown here for readability). We then
%build a random field model on these detection windows by connecting them with
%edges. The edge weights are computed based on the characteristics of the two
%windows. Edges shown in green indicate that the two windows it connects have a
%high probability of occurring together. Edges shown in red connect two windows
%that are unlikely to be characters following one another. A edge shown in red
%forces one of the two windows to take the $\epsilon$ label,~\textit{i.e.}
%removes it from the candidate character set. Based on these edges and the
%\textsc{svm} scores for each window, we infer the character classes of each
%window as well the word, which is indicated by the green edge path. (Best
%viewed in colour.)}
%\label{fig:example}
%\end{figure}
We solve the problem of minimizing the energy function (\ref{eqn:energy}) on a
corresponding graph, where each random variable is represented as a node in the
graph. We begin by ordering the character windows based on their horizontal
location in the image, and add one node each for every window sequentially from
left to right. The nodes are then connected by edges. Since it is not natural
for a window on the extreme left to be strongly related to another window on
the extreme right, we only connect windows which are close to each other. The
intuition behind close-proximity windows is that they could represent
detections of two separate characters. As we will see later, the edges are used
to encode the language model as top-down cues. Such pairwise language priors
alone may not be sufficient in some cases, for example, when an image-specific
lexicon is unavailable. Thus, we also integrate higher order language priors in
the form of $n$-grams computed from the English dictionary by adding an
auxiliary node connecting a set of $n$ character detection nodes.

Each (non-auxiliary) node in the graph takes one label from the label set
$\mathcal{L} = \{l_{1}, l_{2}, \ldots, l_{k}\} \cup \epsilon$. Recall that each
$l_{u}$ is an English character or digit, and the null label $\epsilon$ is used
to discard false windows that represent background or parts of characters. The
cost associated with this label assignment is known as the unary cost. The cost
for two neighbouring nodes taking labels $l_{u}$ and $l_{v}$ is known as the
pairwise cost. This cost is computed from bigram scores of character
pairs in the English dictionary or an image-specific lexicon. The auxiliary
nodes in the graph take labels from the extended label set $\mathcal{L}_{e}$.
Each element of $\mathcal{L}_{e}$ represents one of the $n$-grams present in
the dictionary and an additional label to assign a constant (high) cost to all
$n$-grams that are not in the dictionary. The proposed model is illustrated in
Fig.~\ref{fig:pgm}, where we show a CRF of order four as an example. Once the
graph is constructed, we compute its corresponding cost functions as follows.

\subsubsection{Unary cost}
The unary cost of a node taking a character label is determined by the SVM
confidence scores. The unary term $\psi_{1}$, which denotes the cost of a node
$x_i$ taking label $l_u$, is defined as:
\begin{equation}
\psi_{1}(x_{i}=l_{u}) = 1 - p(l_{u} | x_{i}),
\label{eq:unary1} 
\end{equation}
where $p(l_{u} | x_{i})$ is the SVM score of character class $l_{u}$ for node
$x_{i}$, normalized with Platt's method~\cite{Platt99SVM}. The cost of $x_{i}$
taking the null label $\epsilon$ is given by:
\begin{equation}
\psi_{1}(x_{i}=\epsilon) = \max_{u} p(l_{u} | x_{i}) \exp\left(-\frac{(\mu_{a_u} - a_{i})^{2}}{\sigma_{a_u}^2}\right),
\label{eq:unary2}
\end{equation}
where $a_{i}$ is the aspect ratio of the window corresponding to node $x_{i}$,
$\mu_{a_u}$ and $\sigma_{a_u}$ are the mean and variance of the aspect ratio
respectively of the character $l_u$, computed from the
training data. The intuition behind this cost function is that, for taking a
character label, the detected window should have a high classifier confidence
and its aspect ratio should agree with that of the corresponding character in
the training data.

\subsubsection{Pairwise cost}
The pairwise cost of two neighbouring nodes $x_{i}$ and $x_{j}$ taking a pair
of labels $l_{u}$ and $l_{v}$ respectively is determined by the cost of
their joint occurrence in the dictionary. This cost $\psi_2$ is given by:
\begin{equation}
\psi_{2}(x_{i}=l_{u},x_{j}=l_{v}) = \lambda_{\noit{l}} \exp(- \beta p(l_{u},l_{v})),
\label{eq:pair1} 
\end{equation}
where $p(l_{u},l_{v})$ is the score determining the likelihood of the pair
$l_{u}$ and $l_{v}$ occurring together in the dictionary. The parameters
$\lambda_\noit{l}$ and $\beta$ are set empirically as $\lambda_\noit{l}=2$ and
$\beta=50$ in all our experiments. The score $p(l_{u},l_{v})$ is commonly
computed from joint occurrences of characters in the
lexicon~\cite{thillou2005,elagouni2011,elagouni2012combining,SmithR11}. This
prior is effective when the lexicon size is small, but it is less so as the
lexicon increases in size. Furthermore, it fails to capture the
location-specific information of pairs of characters. As a toy example,
consider a lexicon with only two words CVPR and ICPR. Here, the character pair
{(P,R)} is more likely to occur at the end of the word, but a standard bigram
prior model does not incorporate this location-specific information.
%Further, since numbers are not part of lexicon (especially in case of
%open vocabulary recognition), but often appear as scene text, e.g., house
%numbers, we define joint probabilities of digit pairs as a small non-zero
%value, to avoid any bias towards non-digit word recognition.

To overcome the lack of location-specific information, we devise a
node-specific pairwise cost by adapting~\cite{Riseman} to the scene text
recognition problem.
%Node-specific $n$-grams have been used for contextual post-processing in the past~\cite{Riseman}.
We divide a given word image into $T$ parts, where $T$ is an estimate of the
number of characters in the image. This estimate $T$ is given by the image
width divided by the average character window width, with the average computed
over all the detected characters in the image. To determine the pairwise cost
involving windows in the $t$ th image part, we define a region of interest
(ROI) which includes the two adjacent parts $t-1$, $t+1$, in addition to $t$.
With this, we do a ROI based search in the lexicon. In other words, we consider
all the character pairs involving characters in locations $t-1$, $t$ and $t+1$
in all the lexicon words to compute the likelihood of a pair occurring
together. Note that the extreme cases (involving the leftmost and rightmost
character in the lexicon word) are treated appropriately by considering only
one of the two pairs.
%We divide each word image into $T$ parts, where $T$ is an approximate estimate
%of number of characters in the word image, and is computed as word image width
%divided by average character window width. Here average is computed over all
%the detected character windows on a given word image. Now, for the computation
%of pairwise cost for edges starting in the $t$ th part of the word image we
%define a region of interest (ROI) as $(t-1,t+1)$ th pair of characters in the
%lexicon word. In this process boundary conditions were taken care, e.g., if a
%lexicon word has only $t$ pair of characters, then obviously ROI becomes
%$(t-1,t)$ th pairs of characters in the lexicon word. In other words, we do a
%region of interest (ROI) based search in the lexicon. The ROI is determined
%based on the spatial position of a detected window in the word, e.g., if two
%windows are on the left most side then only the first couple of characters of
%lexicons are considered for calculating the pairwise term between windows.

This pairwise cost using the node-specific prior is given by:
\begin{equation}
\psi_{2}(x_{i}=l_{u},x_{j}=l_{v}) = \left\{
\begin{array}{ll}
0 & \text{if $(l_{u},l_{v}) \in$ \textsc{roi},} \\
\lambda_\noit{l} & \mbox{otherwise}. \end{array}
\right.
\label{eq:lexicon2}
\end{equation}
We evaluated our approach with both the pairwise terms (\ref{eq:pair1}) and
(\ref{eq:lexicon2}), and found that the node-specific prior (\ref{eq:lexicon2})
achieves better performance. The cost of nodes $x_{i}$ and $x_{j}$ taking label
$l_{u}$ and $\epsilon$ respectively is defined as:
\begin{equation}
\psi_2(x_{i} = l_{u},x_{j} = \epsilon) = \lambda_\noit{o} \exp(-\beta(1-O(x_{i},x_{j}))^{2}),
\label{eq:pair2}
\end{equation}
where $O(x_{i},x_{j})$ is the overlap fraction between windows corresponding to
the nodes $x_{i}$ and $x_{j}$. The pairwise cost $\psi_2(x_{i} = \epsilon,
x_{j} = l_{u})$ is defined similarly. The parameters are set empirically as
$\lambda_\noit{o}=2$ and $\beta=50$ in our experiments. This cost ensures that when two character windows overlap significantly, only one of them are assigned a
character/digit label in order to avoid parts of characters being labelled.

\subsubsection{Higher order cost}
Let us consider a CRF of order $n = 3$ as an example to understand this cost.
An auxiliary node corresponding to every clique of size $3$ is added to
represent this third order cost in the graph. The higher order cost is then
decomposed into unary and pairwise terms with respect to this node, similar
to~\cite{luborUAI10}. Each auxiliary node in the graph takes one of the labels
from the extended label set $\{L_{1},L_{2},\ldots,L_{M}\} \cup L_{M+1}$, where
labels $L_{1} \dots L_{M}$ represent all the trigrams in the dictionary. The
additional label $L_{M+1}$ denotes all those trigrams which are absent in the
dictionary. The unary cost $\psi^a_{1}$ for an auxiliary variable $y_i$ taking
label $L_m$ is:
\begin{equation}
\psi^a_{1}(y_{i}=L_{m}) = \lambda_\noit{a} \exp(-\beta P(L_{m})),
\label{eq:auxiuni}
\end{equation}
where $\lambda_\noit{a}$ is a constant. We set $\lambda_\noit{a}=5$
empirically, in all our experiments, unless stated otherwise. The parameter
$\beta$ controls penalty between dictionary and non-dictionary $n$-grams, and
is empirically set to $50$. The score $P(L_m)$ denotes the likelihood of
trigram $L_m$ in the English, and is further described in
Section~\ref{sec:langPriors}. The pairwise cost between the auxiliary node
$y_i$ taking a label $L_{m} = l_{u} l_{v} l_{w}$ and the left-most
non-auxiliary node in the clique, $x_{i}$, taking a label $l_{r}$ is given by:
\begin{equation}
\psi^a_{2}(y_{i}=L_{m},x_{i} = l_{r}) = \left\{
\begin{array}{ll}
0 & \text{if $r=u$}\\
0 & \text{if $l_r=\epsilon$}\\
\lambda_{\noit{b}} & \mbox{otherwise}, \end{array}
\right.
\label{eq:auxipair}
\end{equation}
where $\lambda_{\noit{b}}$ penalizes a disagreement between the auxiliary and
non-auxiliary nodes, and is empirically set to $1$. The other two pairwise
terms for the second and third nodes are defined similarly. Note that when one
or more $x_i$'s take null label, the corresponding pairwise term(s) between
$x_i$(s) and the auxiliary node are set to $0$.

\subsubsection{Computing language priors}
\label{sec:langPriors}
We compute $n$-gram based priors from the lexicon (or dictionary) and then
adapt standard techniques for smoothing these
scores~\cite{thillou2005,katz,Goodman01abit} to the open and closed
vocabulary cases.

Our method uses the score denoting the likelihood of joint occurrence of pair
of labels $l_u$ and $l_v$ represented as $P(l_u,l_v)$, triplets of labels
$l_u$, $l_v$ and $l_w$ denoted by $P(l_u,l_v,l_w)$ and even higher order (e.g.,
fourth order). Let $C(l_u)$ denote the number of occurrences of $l_u$,
$C(l_u,l_v)$ be the number of joint occurrences of $l_u$ and $l_v$ next to each
other, and similarly $C(l_u,l_v,l_w)$ is the number of joint occurrences of all
three labels $l_u, l_v, l_w$ next to each other. The smoothed
scores~\cite{katz} $P(l_u,l_v)$ and $P(l_u,l_v,l_w)$ are now:
\begin{equation}
P(l_u,l_v) = \left\{
\begin{array}{ll}
0.4 & \text{if $l_u, l_v$ are digits},\\
\frac{C(l_u,l_v)}{C(l_v)} & \text{if $C(l_u,l_v) > 0$},\\
\alpha_{l_u}P(l_v) & \mbox{otherwise}, \end{array}
\right.
\end{equation}

\begin{equation}
P(l_u,l_v,l_w) = \left\{
\begin{array}{ll}
0.4 & \text{if $l_u, l_v, l_w$ are digits},\\
\frac{C(l_u,l_v,l_w)}{C(l_v,l_w)} & \text{if $C(l_u,l_v,l_w) > 0$},\\
\alpha_{l_u}P(l_v,l_w) & \text{else if $C(l_u,l_v) > 0$},\\
\alpha_{l_u,l_v} P(l_w) & \mbox{otherwise}, \end{array}
\right.
\end{equation}
Image-specific lexicons (small or medium) are used in the closed vocabulary
setting, while in the open vocabulary case we use a lexicon containing half a
million words (henceforth referred to as large lexicon) provided
by~\cite{WeinmanLH09} to compute these scores. The parameters
$\alpha_{l_u}$ and $\alpha_{l_u,l_v}$ are learnt on the large lexicon using
SRILM toolbox.\footnote{Available at:
\url{http://www.speech.sri.com/projects/srilm/}} They determine the low
score values for $n$-grams not present in the lexicon. We assign a
constant value ($0.4$) when the labels are digits, which do not occur in the
large lexicon. 
\subsubsection{Inference}
Having computed the unary, pairwise and higher order terms, we use the
sequential tree-reweighted message passing (TRW-S)
algorithm~\cite{Kolmogorov06} to minimize the energy function. The TRW-S
algorithm maximizes a concave lower bound of the energy. It begins by
considering a set of trees from the random field, and computes probability
distributions over each tree. These distributions are then used to reweight the
messages being passed during loopy belief propagation~\cite{Pearl88} on each
tree. The algorithm terminates when the lower bound cannot be increased
further, or the maximum number of iterations has been reached.

In summary, given an image containing a word, we: (i) locate the potential
characters in it with a character detection scheme, (ii) define a random field
over all these potential characters, (iii) compute the language priors and
integrate them into the random field model, and then (iv) infer the most likely
word by minimizing the energy function corresponding to the random field.

\begin{table}[!t]
\renewcommand{\arraystretch}{1.3}
\caption{Our IIIT 5K-word dataset contains a few less challenging (Easy) and
many very challenging (Hard) images. To present analysis of the dataset, we
manually divided the words in the training and test sets into {\it easy} and
{\it hard} categories based on their visual appearance. The recognition
accuracy of a state-of-the-art commercial OCR -- ABBYY9.0 -- for this dataset
is shown in the last column. Here we also show the total number of characters,
whose annotations are also provided, in the dataset.}
\centering
\begin{tabular}{|l|c|c|c|}
\hline
 &\multicolumn{3}{c|}{{Training Set}}\\
\cline{2-4}
 &{$\#$words}&{$\#$characters}&{ABBYY9.0(\%)}\\
\hline
{Easy} & 658 & - & 44.98 \\
{Hard} & 1342 & - & 16.57 \\
{Total}& 2000 & 9658 & 20.25\\ 
%ABBYY9.0(+ binarization) & 43.74 & 24.37 & 30.08 &42.51&18.45&24.33\\
\hline\hline
&\multicolumn{3}{c|}{{Test Set}}\\
\cline{2-4}
&{$\#$words}&{$\#$characters}&{ABBYY9.0(\%)}\\
\hline
{Easy}&734 & - & 44.96\\
{Hard}& 2266 & - & 5.00 \\
{Total}& 3000 & 15269 & 14.60 \\
\hline
\end{tabular}
\label{tab:ourdataset}
\end{table}
\section{Datasets and Evaluation Protocols}
\label{datasets}
Several public benchmark datasets for scene text understanding have been
released in recent years. ICDAR~\cite{ICDAR} and Street View Text
(SVT)~\cite{SVT} datasets are two of the initial datasets for this problem.
They both contain data for text localization, cropped word recognition and
isolated character recognition tasks. In this paper we use the cropped word
recognition part from these datasets. Although these datasets have served well
in building interest in the scene text understanding problem, they are limited
by their size of a few hundred images. To address this issue, we introduced the
IIIT 5K-word dataset~\cite{MishraBMVC12}, containing a diverse set of 5000
words. Here, we provide details of all these datasets and the evaluation
protocol.

\paragraph{SVT} The street view text (SVT) dataset contains images taken from
Google Street View. As noted in~\cite{WangB10}, most of the images come from
business signage and exhibit a high degree of variability in appearance and
resolution. The dataset is divided into SVT-spot and SVT-word, meant for the
tasks of locating and recognizing words respectively. We use the SVT-word
dataset, which contains 647 word images.

Our basic unit of recognition is a character, which needs to be localized
before classification. Failing to detect characters will result in poorer word
recognition, making it a critical component of our framework. To quantitatively
measure the accuracy of the character detection module, we created ground truth
data for characters in the SVT-word dataset. This ground truth dataset contains
around 4000 characters of 52 classes, and is referred to as as SVT-char, which
is available for download~\cite{project}.
\setlength{\tabcolsep}{2pt}
\begin{table}[!t]
\centering
\caption{Analysis of the IIIT 5K-word dataset. We show the percentage of
non-dictionary words (Non-dict.), including digits, and the percentage of words
containing only digits (Digits) in the first two rows. We also show the
percentage of words that are composed from valid English trigrams (Dict.\
3-grams), four-grams (Dict.\ 4-grams) and five-grams (Dict.\ 5-grams) in the
last three rows. These statistics are computed using the large lexicon.}
\begin{tabular}{|l|c|c|}
\hline
                 & IIIT 5K train & IIIT 5K test \\
\hline
Non-dict.\ words &23.65 &  22.03\\
Digits           &11.05 &  7.97\\
Dict.\ 3-grams   &90.27 &  88.05\\
Dict.\ 4-grams   &81.40 &  79.27\\
Dict.\ 5-grams   &68.92 &  62.48\\ 
\hline
\end{tabular}
\label{tab:langModel}
\end{table}
\paragraph{ICDAR 2003 dataset} The ICDAR 2003 dataset was originally created
for text detection, cropped character classification, cropped and full image
word recognition, and other tasks in document analysis~\cite{ICDAR}. We used
the part corresponding to the cropped word recognition called robust word
recognition. Following the protocol of~\cite{WangB11}, we ignore words with
less than two characters or with non-alphanumeric characters, which results in
859 words overall. For subsequent discussion we refer to this dataset as
ICDAR(50) for the image-specific lexicon-driven case (closed vocabulary), and
ICDAR 2003 when this lexicon is unavailable (open vocabulary case).

\paragraph{ICDAR 2011/2013 datasets} These datasets were introduced as part of
the ICDAR robust reading competitions~\cite{ICDAR11Comp,ICDAR13Comp}. They
contain 1189 and 1095 word images respectively. We show case-sensitive open
vocabulary results on both these datasets. Also, following the ICDAR
competition evaluation protocol, we do not exclude words containing special
characters (such as \&, :), and report results on the entire dataset.

\paragraph{IIIT 5K-word dataset} The IIIT 5K-word
dataset~\cite{MishraBMVC12,project} contains both scene text and born-digital
images. Born-digital images---category of images which has gained interest in
ICDAR 2011 competitions~\cite{ICDAR11Comp}---are inherently low-resolution,
made for online transmission, and have a variety of font sizes and styles. This
dataset is not only much larger than SVT and the ICDAR datasets, but also more
challenging. All the images were harvested through Google image search. Query
words like billboard, signboard, house number, house name plate, movie poster
were used to collect images. The text in the images was manually annotated with
bounding boxes and their corresponding ground truth words. The IIIT 5K-word
dataset contains in all 1120 scene images and 5000 word images. We split it
into a training set of 380 scene images and 2000 word images, and a test set of
740 scene images and 3000 word images. To analyze the difficulty of the IIIT
5K-word dataset, we manually divided the words in the training and test sets
into {\it easy} and {\it hard} categories based on their visual appearance.
An annotation team consisting of three people have done three independent splits. Each word is then tagged as either being easy or hard by taking a majority vote. This 
split is available on our project page~\cite{project}. Table~\ref{tab:ourdataset} shows these splits in detail. We observe that a commercial OCR performs poorly on both the train and test splits. Furthermore, to evaluate components like character detection and recognition, we also
provide annotated character bounding boxes. It should be noted that around 22\%
of the words in this dataset are not in the English dictionary, e.g., proper
nouns, house numbers, alphanumeric words. This makes this dataset suitable for
open vocabulary cropped word recognition. We show an analysis of dictionary and
non-dictionary words in Table~\ref{tab:langModel}.
\paragraph{Evaluation protocol} We evaluate the word recognition accuracy in
two settings: closed and open vocabulary. Following previous
work~\cite{WangB11,shiCVPR13,MishraBMVC12}, we evaluate case-insensitive word
recognition on SVT, ICDAR 2003, IIIT 5K-word, and case-sensitive word
recognition on ICDAR 2011 and ICDAR 2013. For the closed vocabulary recognition
case, we perform a minimum edit distance correction, since the ground truth
word belongs to the image-specific lexicon. On the other hand, in the case of
open vocabulary recognition, where the ground truth word may or may not belong
to the large lexicon, we do not perform edit distance based correction. We
perform many of our analyses on the IIIT 5K-word dataset, unless otherwise
stated, since it is the largest dataset for this task, and also comes with
character bounding box annotations.
\begin{table*}[!t]
\centering
\caption{Character classification accuracy (in \%). A smart choice of features,
training examples and classifier is key to improving character classification.
We enrich the training set by including many affine transformed (AT) versions
of the original training data from ICDAR and Chars74K (c74k). The three variants of our approach (H-13, H-31 and H-36) show noticeable improvement over several methods.
%Here H-36 is 36-dimensional HOG proposed in~\cite{DalalT05}, while H-13 and H-31 are
%proposed in~\cite{Felzen10}. 
The character classification results shown here
are case sensitive (all rows except the last two). It is to be noted
that~\cite{CamposBV09} only uses 15 training samples per class. The last two
rows show a case insensitive (CI) evaluation. $*$We do not evaluate the
convolutional neural network classifier in~\cite{AZ14} (CNN feat+classifier) on
the c74K dataset, since the entire dataset was used to train the network.}
\begin{tabular}{|l|c|c|c|c|c|}
\hline
{Method} & {SVT} & {ICDAR} & {c74K} & {IIIT 5K} & {Time} \\
\hline
Exempler SVM~\cite{SheshadriD12}&-&71&-&-&-\\
Elagouni {\it et al}.~\cite{elagouni2012combining}&-&70&-&-&-\\
Coates {\it et al}.~\cite{CoatesCCSSWWN11}&-&{82}&-&-&-\\
FERNS~\cite{WangB11}& - & 52 & 47 & - & -\\
RBF~\cite{MishraCVPR12} & 62 & 62 & 64 & 61 &3ms  \\
MKL+RBF~\cite{CamposBV09} & - & - & 57 & - &11ms \\
H-36+AT+Linear &69 &73&68 &{66}& 2ms  \\
H-31+AT+Linear & 64 & 73 & 67 &63 &1.8ms \\
H-13+AT+Linear & 65 & 72 & 66 &64&0.8ms \\
\hline\hline
H-36+AT+Linear (CI) & 75 & 77 & {79} & 75 &0.8ms\\
CNN feat+classifier~\cite{AZ14} (CI) & {83} & {86} & $*$ & {85} & 1ms \\
\hline
\end{tabular}
\label{tab:charClassification}
\end{table*}

\section{Experiments}
\label{sec:expts}
Given an image region containing text, cropped from a street scene, our task is
to recognize the word it contains. In the process, we develop several
components (such as a character recognizer) and also evaluate them to justify
our choices. The proposed method is evaluated in two settings, namely, closed
vocabulary (with an image-specific lexicon) and open vocabulary (using an
English dictionary for the language model). We compare our results with the best-performing recent methods for these two cases. For baseline comparisons
we choose commercial OCR namely ABBYY~\cite{ABBYY} and a public implementation
of a recent method~\cite{Gomez14} in combination with an open source OCR.
\subsection{Character Classifier}
\label{subsec:charclassif}
We use the training sets of ICDAR 2003 character~\cite{ICDAR} and
Chars74K~\cite{CamposBV09} datasets to train the character classifiers. This
training set is augmented with $48\times48$ patches harvested from scene
images, with buildings, sky, road and cars, which do not contain text, as
additional negative training examples. We then apply affine transformations to
all the character images, resize them to $48\times48$, and compute HOG
features. Three variations (13, 31 and 36-dimensional) of HOG were analyzed
(see Table~\ref{tab:charClassification}). We then use an explicit feature
map~\cite{VedaldiPAMI12} and the $\chi^2$ kernel to learn the SVM classifier.
The SVM parameters are estimated by cross-validating on a validation set. The
explicit feature map not only allows a significant reduction in classification
time, compared to non-linear kernels like RBF, but also achieves a good
performance.

The two main differences from our previous work~\cite{MishraCVPR12} in the
design of the character classifier are: (i) enriching the training set, and
(ii) using an explicit feature map and a linear kernel (instead of RBF).
Table~\ref{tab:charClassification} compares our character classification
performance
with~\cite{WangB11,CamposBV09,MishraCVPR12,SheshadriD12,CoatesCCSSWWN11,elagouni2012combining}
on several test sets. We achieve at least 4\% improvement over our
previous work (RBF~\cite{MishraCVPR12}) on all the datasets, and also perform
better than~\cite{WangB11,CamposBV09}. We are also comparable to a few other
recent methods~\cite{elagouni2012combining,SheshadriD12}, which show a limited
evaluation on the ICDAR 2003 dataset. Following an evaluation insensitive to
case (as done in a few benchmarks, e.g.,~\cite{AZ14,shiCVPR13},
we obtain 77\% on ICDAR 2003, 75\% on SVT-char, 79\% on Chars74K, and 75\% on
IIIT 5K-word. It should be noted that feature learning methods based on
convolutional neural networks, e.g.,~\cite{CoatesCCSSWWN11,AZ14}, show an
excellent performance. This inspired us to integrate them into our framework.
We used publicly available features~\cite{AZ14}. This will be further discussed
in Section~\ref{sec:wordReco}. We could not compare with other related recent
methods~\cite{photoOCR,weinman2013toward} since they did not report isolated
character classification accuracy.

In terms of computation time, linear SVMs trained with HOG-13 features
outperform others, but since our main focus is on word recognition performance,
we use the most accurate combination, i.e., linear SVMs with HOG-36. We
observed that this smart selection of training data and features not only
improves character recognition accuracy but also improves the second and third
best predictions for characters.
\subsection{Character Detection}
\label{subsec:chars}
Sliding window based character detection is an important component of our
framework, since our random field model is defined on these detections. 
We use windows of aspect ratio ranging from 0.1 to 2.5 for sliding window and at every possible location of the sliding window, we evaluate a character
classifier. This provides the likelihood of the window containing the
respective character. We pruned some of the windows based on their aspect
ratio, and then used the goodness measure~(\ref{eq:gs}) to discard the windows
with a score less than $0.1$ (refer Section~\ref{sec:charDet}).
Character-specific NMS is done on the remaining windows with an overlap
threshold of $40\%$, i.e., if two detections have more than 40\% overlap and
represent the same character class, we suppress the weaker detection.  We
evaluated the character detection results with the intersection over union
measure and a threshold of 50\%, following ICDAR 2003~\cite{ICDAR} and
PASCAL-VOC~\cite{Everingham10} evaluation protocol. Our sliding window approach
achieves recall of 80\% on the IIIT 5K-word dataset, significantly better than
using a binarization scheme for detecting characters and also superior to
techniques like MSER~\cite{mser} and CSER~\cite{Gomez14} (see
Table~\ref{tab:binResults} and Section~\ref{binMethods}).
% We operate in such a high recall mode, 
%since our model is capable of suppressing false positive windows.
%\begin{table}[!t]
%\renewcommand{\arraystretch}{1.3}
%\centering
%\caption{Evaluation of the performance of our sliding window approach. We
%use the intersection over union measure~\cite{ICDAR,Everingham10}
%to determine whether a detection has been retrieved or not.
%Note that high character detection recall is important for our word recognition performance.}
%\begin{tabular}{|l|c|}
%\hline
%\textbf{Method} & \textbf{Recall} \\
%\hline
%No {\sc nms} &0.99\\
%Char. Specific {\sc nms} + pruning& 0.80 \\
%\hline
%\end{tabular}
%\label{tab:det}
%\end{table}
\begin{table}[!t]
\centering
\caption{Word recognition accuracy (in~\%): closed vocabulary setting. We
present results of our proposed higher order model (``This work'') with HOG as
well as CNN features. See text for details.}
\begin{tabular}{ |l|c| }
  \hline
   {Method} & {Accuracy}\\
  \hline \hline
  \multicolumn{2}{|c|}{{ICDAR 2003 (50) dataset}} \\
  \hline \hline
~~~Baseline (ABBYY)~\cite{ABBYY} & 56.04 \\
~~~Baseline (CSER+tesseract)~\cite{Gomez14} & 57.27 \\
~~~Novikova {\it et al}.~\cite{pushmeetLexi}&82.80\\
~~~Our Holistic recognition~\cite{GoelICDAR13}&89.69\\
~~\textit{Deep learning approaches} & \\
~~~Wang {\it et al}.~\cite{ngICPR12}&90.00\\
~~~Deep features~\cite{AZ14}&{96.20}\\
~~\textit{Other energy min.\ approaches} & \\
~~~{PLEX}~\cite{WangB11} & 72.00 \\
~~~Shi {\it et al.}~\cite{shiCVPR13}&87.04\\
~~\textit{Our variants:} &\\
~~~Pairwise CRF~\cite{MishraCVPR12}& 81.74\\
~~~Higher order [This work, HOG] & {84.07}  \\
~~~Higher order [This work, CNN] & {88.02} \\
  \hline \hline
\multicolumn{2}{|c|}{{SVT-Word dataset}} \\
  \hline \hline
~~~Baseline(ABBYY)~\cite{ABBYY} & 35.00  \\
~~~Baseline (CSER+tesseract)~\cite{Gomez14} &37.71\\
~~~Novikova {\it et al.}~\cite{pushmeetLexi}&72.90\\
~~~Our Holistic recognition~\cite{GoelICDAR13}&77.28\\
~~\textit{Deep learning approaches} & \\
~~~Wang {\it et al.}~\cite{ngICPR12}&70.00\\
~~~PhotoOCR~\cite{photoOCR}&{90.39}\\
~~~Deep features~\cite{AZ14}& 86.10\\
~~\textit{Other energy min.\ approaches} & \\
~~~{PICT}~\cite{WangB10} & 59.00\\
~~~{PLEX}~\cite{WangB11} & 57.00 \\
~~~Shi {\it et al.}~\cite{shiCVPR13}&73.51\\
~~~Weinman {\it et al.}~\cite{weinman2013toward}&78.05\\
~~\textit{Our variants:} & \\
~~~Pairwise {CRF}~\cite{MishraCVPR12}& 73.26 \\
~~~Higher order [This work, HOG] & 75.27 \\
~~~Higher order [This work, CNN] & 78.21 \\
  \hline \hline
\multicolumn{2}{|c|}{{IIIT 5K-Word (Small)}} \\
  \hline \hline
~~~Baseline(ABBYY)~\cite{ABBYY} & 24.50\\
~~~Baseline (CSER+tesseract)~\cite{Gomez14} & 33.07\\
~~~Rodriguez \& Perronnin~\cite{JoseBMVC13}&76.10\\
~~~Strokelets~\cite{stroklets}&{80.20}\\
~~\textit{Our variants:} & \\
~~~Pairwise {CRF}~\cite{MishraCVPR12}& 66.13\\
~~~Higher order [This work, HOG] & 71.80  \\
~~~Higher order [This work, CNN] & {78.07}  \\
\hline
\end{tabular}
\label{tab:smallLexiRes}
\end{table}
\subsection{Word Recognition}
\label{sec:wordReco}
\paragraph{Closed vocabulary recognition}
The results of the proposed CRF model in closed vocabulary setting are
presented in Table~\ref{tab:smallLexiRes}. We compare our method with many
recent works for this task. To compute the language priors we use lexicons
provided by authors of~\cite{WangB11} for SVT and ICDAR(50). The image-specific
lexicon for every word in the IIIT 5K-word dataset was developed following the
method described in~\cite{WangB11}. These lexicons contain the ground truth
word and a set of distractors obtained from randomly chosen words (from all the
ground truth words in the dataset). We used a CRF with higher order term
($n$=4), and similar to other approaches, applied edit distance based
correction after inference. The constant $\lambda_\noit{a}$ in
(\ref{eq:auxiuni}) to 1, given the small size of the lexicon.

The gain in accuracy over our previous work~\cite{MishraCVPR12}, seen in
Table~\ref{tab:smallLexiRes}, can be attributed to the higher order CRF and an
improved character classifier. The character classifier uses: (i) enriched
training data, and (ii) an explicit feature map, to achieve about 5\% gain (see
Section~\ref{subsec:charclassif} for details). Other methods, in particular,
our previous work on holistic word recognition~\cite{GoelICDAR13}, label
embedding~\cite{JoseBMVC13} achieve a reasonably good performance, but are
restricted to the closed vocabulary setting, and their extension to more
general settings, such as the open vocabulary case, is unclear. Methods
published since our original work~\cite{MishraCVPR12}, such
as~\cite{weinman2013toward,shiCVPR13}, also perform well. Very recently,
methods based on convolutional neural networks~\cite{photoOCR,AZ14} have shown
very impressive results for this problem. It should be noted that such methods
are typically trained on much larger datasets, for example, 10M compared to
0.1M typically used in state-of-the-art methods, which are not publicly
available~\cite{photoOCR}. Inspired by these successes, we use a CNN
classifier~\cite{AZ14} to recognize characters, instead of our SVM classifier
based on HOG features (see Sec.~\ref{sec:charDet}). We show results with this CNN
classifier on SVT, ICDAR 2003 and IIIT-5K word datasets in
Table~\ref{tab:smallLexiRes} and observe significant improvement in accuracy,
showing its complementary nature to our energy based method. However, there remains a difference in performance between the deep feature based method~\cite{AZ14} and [This work, CNN]. This is primarily due to use of CNN features for learning classifiers for individual character as well as bi-grams in~\cite{AZ14}. In contrast, our method only uses the pre-trained character classifier provided by~\cite{AZ14}. Nevertheless, the improvement observed over [This work, HOG] does show the complementary nature of the two approaches, and integrating the two further would be an interesting avenue for future research.

\paragraph{Open vocabulary recognition}
In this setting we use a lexicon of 0.5 million words from~\cite{WeinmanLH09}
instead of image-specific lexicons to compute the language priors. Many character pairs are equally likely in such a large lexicon, thereby rendering pairwise priors is less effective than in the case of a small lexicon. We use priors of order four to address this (see also analysis on the CRF order in Section~\ref{sec:analysis}). Results on
various datasets in this setting are shown in Table~\ref{tab:largeLexiRes}.  We
compare our method with recent work by Feild and Miller~\cite{fieldICDAR13} on
the ICDAR 2003 dataset, where our method with HOG features shows a comparable
performance. Note that~\cite{fieldICDAR13} additionally uses web-based
corrections, unlike our method, where the results are obtained directly by
performing inference on the higher order {CRF} model. On the ICDAR 2011 and
2013 datasets we compare our method with the top performers from the respective
competitions. Our method outperforms the ICDAR 2011 robust reading competition
winner (TH-OCR method) method by 17\%. This performance is also better than
a recently published work from 2014 by Weinman {\it et
al.}~\cite{weinman2013toward}. On the ICDAR 2013 dataset, the proposed higher
order model is significantly better than the baseline and is in the top-5
performers among the competition entries. The winner of this competition
(PhotoOCR) uses a large proprietary training dataset, which is unavailable
publicly, making it infeasible to do a fair comparison. Other methods
(NESP~\cite{NESP}, MAPS~\cite{MAPS}, PLT~\cite{PLT}) use many preprocessing
techniques, followed by off-the-self OCR. Such preprocessing techniques are
highly dataset dependent and may not generalize easily to all the challenging
datasets we use. Despite the lack of these preprocessing steps, our method
shows a comparable performance. On the IIIT 5K-word dataset, which is large
(three times the size of ICDAR 2013 dataset) and challenging, the only
published result to our knowledge
%\footnote{Another recent method~\cite{AlmazanGFV14} shows retrieval results on IIIT 5K-word, but does not report recognition results in the open vocabulary setting.}
is Strokelets~\cite{stroklets} from CVPR 2014. Our method performs 7\% better
than Strokelets. Using CNN features instead of HOG further improves our word
recognition accuracy, as shown in Table~\ref{tab:largeLexiRes}.

The main focus of this work is on evaluating datasets containing scene text images or a mixture of scene text and born-digital images. Nevertheless, we also tested our method on the born-digital image dataset from the recent ICDAR 2013 competition. Our approach with pre-trained CNN features achieves 78\% accuracy on this dataset, which is comparable to other top performers (80.40\%, 80.26\%, 79.40\%), and lower than PhotoOCR (82\%), the competition winner using an end-to-end deep learning approach.
 
To sum up, our proposed method performs well consistently on several popular
scene text datasets. Fig.~\ref{fig:word_results} shows the qualitative
performance of the proposed method on a few sample images. The higher order CRF
outperforms the unary and pairwise CRFs. This is intuitive due to the better
expressiveness of the higher order potentials. One of the failure cases is
shown in the last row in Fig.~\ref{fig:word_results}, where the higher order
potential is computed from a lexicon which does not have sufficient examples to
handle alphanumeric words.
\begin{table}[!t]
\renewcommand{\arraystretch}{0.9}
\centering
\caption{\small{Word recognition accuracy (in \%): open vocabulary setting. The
results of our proposed higher order model (``This work'') with HOG as well as
CNN features are presented here. Since the network used here to compute CNN
features, i.e.~\cite{AZ14}, is learnt on data from several sources (e.g., ICDAR
2013), we evaluated with CNN features only on ICDAR 2003 and IIIT-5K word
datasets, as recommended by the authors. Note that we also compare with top performers (as given in~\cite{ICDAR11Comp,ICDAR13Comp}) in the ICDAR 2011 and 2013 robust
reading competitions. We follow standard protocols for evaluation -- case
sensitive on ICDAR 2011 and 2013 and case insensitive on ICDAR 2003 and IIIT
5K-Word.}}
\begin{tabular}{ |l|c| }
  \hline
   {Method} & {Accuracy}\\
  \hline \hline
\multicolumn{2}{|c|}{{ICDAR 2003 dataset}} \\
  \hline \hline
~~~Baseline (ABBYY) & 46.51\\
~~~Baseline (CSER+tesseract)~\cite{Gomez14} & 50.99\\
~~~Feild and Miller~\cite{fieldICDAR13} & 62.76\\
~~\textit{Our variants} & \\
%~~Otsu~\cite{otsu} + CRF& 45  \\
%~~MRFbin~\cite{MishraJ11} + CRF& 52   \\
~~~Pairwise~\cite{MishraCVPR12}& 50.99  \\
~~~Higher order [This work, HOG]& {63.02} \\
~~~Higher order [This work, CNN]& {67.67} \\
  \hline \hline
\multicolumn{2}{|c|}{{ICDAR 2011 dataset}} \\
  \hline \hline
~~~Baseline (ABBYY) &46.00\\
~~~Baseline (CSER+tesseract)~\cite{Gomez14} & 51.98 \\
~~~Weinman~\textit{et al.}~\cite{weinman2013toward} & 57.70 \\
~~~Feild and Miller~\cite{fieldICDAR13} & 48.86\\
~~\textit{ICDAR'11 competition}~\cite{ICDAR11Comp} & \\
~~~TH-OCR System & 41.20\\
~~~KAIST AIPR System  & 35.60\\
~~~Neumann's Method & 33.11 \\
~~\textit{Our variants} & \\
%~~Otsu~\cite{otsu} + CRF& 44 \\
%~~MRFbin~\cite{MishraJ11} + CRF&  48  \\
~~~Pairwise~\cite{MishraCVPR12}& 48.11  \\
~~~Higher order [This work, HOG]& {58.03}\\
  \hline \hline
\multicolumn{2}{|c|}{{ICDAR 2013 dataset}} \\
  \hline \hline
~~~Baseline (ABBYY) & 45.30\\
~~~Baseline (CSER+tesseract)~\cite{Gomez14} & 50.26 \\
~~\textit{ICDAR'13 competition}~\cite{ICDAR13Comp}&\\
~~~PhotoOCR~\cite{photoOCR} & {82.83}\\
~~~NESP~\cite{NESP}   & 64.20 \\
~~~MAPS~\cite{MAPS} & 62.74\\
~~~PLT~\cite{PLT} & 62.37\\
~~~PicRead~\cite{pushmeetLexi}  & 57.99\\
~~~POINEER~\cite{WeinmanLH09,weinman2013toward} &53.70\\
~~~Field's Method~\cite{fieldICDAR13} &47.95\\
~~~TextSpotter~\cite{NeumannM10,neumann2012real,NeumannM13} & 26.85\\
~~\textit{Our variants} & \\
%~~Otsu~\cite{otsu} + CRF& 47 \\
%~~MRFbin~\cite{MishraJ11} + CRF&49  \\
~~~Pairwise~\cite{MishraCVPR12}& 49.86   \\
~~~Higher order [This work, HOG]& 60.18  \\
  \hline \hline
\multicolumn{2}{|c|}{{IIIT 5K-Word}} \\
  \hline \hline
~~~Baseline (ABBYY) & 14.60 \\
~~~Baseline (CSER+tesseract)~\cite{Gomez14} & 25.00\\
~~~Stroklets~\cite{stroklets}&38.30\\
~~\textit{Our variants} & \\
%~~Otsu~\cite{otsu} + CRF& 25  \\
%~~MRFbin~\cite{MishraJ11} + CRF& 28   \\
~~~Pairwise~\cite{MishraCVPR12}& 32.00  \\
~~~Higher order [This work, HOG]& {44.50} \\
~~~Higher order [This work, CNN]& {46.73} \\
 \hline
\end{tabular}
\label{tab:largeLexiRes}
\end{table}

\begin{table}[!t]
\caption{Studying the influence of the lexicon size -- small (S), medium (M),
large (L) -- on the IIIT 5K-word dataset in the closed vocabulary setting.
}
\centering
\begin{tabular}{|l|c|c|c|}
\hline
Method & S & M & L \\
\hline
Rodriguez \& Perronnin~\cite{JoseBMVC13} & 76.10 & 57.50 &- \\
Strokelets~\cite{stroklets}& {80.20} & 69.30 & 38.30 \\
Higher order [This work, HOG] & 71.80 & 62.17 & 44.50\\
Higher order [This work, CNN] & 78.07 & {70.13} & {46.73} \\
\hline
\end{tabular}
\label{tab:lexiSize}
\end{table}

\subsection{Further Analysis}
\label{sec:analysis}
\paragraph{Lexicon size} The size of the lexicon plays an important role in the
word recognition performance. With a small-size lexicon, we obtain strong
language priors which help overcome inaccurate character detection and
recognition in the closed vocabulary setting. A small lexicon provides much
stronger priors than the large lexicon in this case, as the performance
degrades with increase in the lexicon size. We show this behaviour on the IIIT
5K-word dataset in Table~\ref{tab:lexiSize} with small (50), medium (1000) and
large (0.5 million) lexicons. We also compare our results with a
state-of-the-art methods~\cite{JoseBMVC13,stroklets}. We observe that~\cite{JoseBMVC13,stroklets} shows better recognition performance with the small lexicon, when we use HOG features, but as the size of the lexicon increases, our method
outperforms~\cite{JoseBMVC13}.

\begin{figure*}[!t]
\centering
\begin{tabular}{cccc}
\textbf{Test Image} &~~~~~\textbf{Unary} &~~~~~\textbf{Pairwise}&~~~~~\textbf{Higher order(=4)}\\
\includegraphics[width=2.5cm,height=0.8cm]{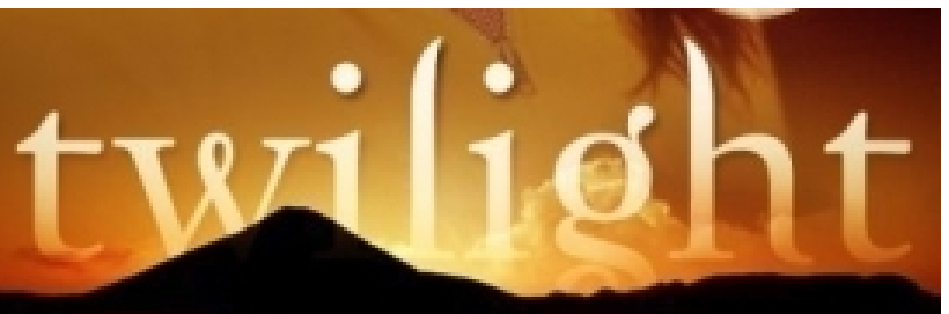} &~~~~~\raisebox{.30cm}{TWI\textcolor{red}{1}I\textcolor{red}{O}HT}  &~~~~~~~~~\raisebox{.30cm}{TWILI\textcolor{red}{O}HT}~~~~~&~~~~~\raisebox{.30cm}{TWILIGHT} \\
\includegraphics[width=2.5cm,height=0.8cm]{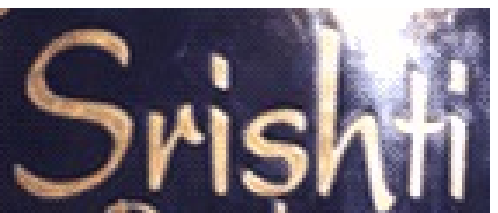} &~~~~~\raisebox{.30cm}{SRIS\textcolor{red}{N}TI}&~~~~~\raisebox{.30cm}{SRIS\textcolor{red}{N}TI}&~~~~~\raisebox{.30cm}{SRISHTI} \\
\includegraphics[width=2.5cm,height=0.8cm]{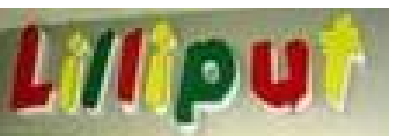} &~~~~~\raisebox{.30cm}{LIIIIPUT}&~~~~~\raisebox{.30cm}{LIIIIPUT}&~~~~~~~\raisebox{.30cm}{LILLIPUT} \\
\includegraphics[width=2.5cm,height=0.8cm]{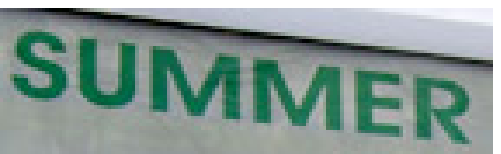} &~~~~~~~\raisebox{.30cm}{\textcolor{red}{E}UMMER} &~~~~~~~\raisebox{.30cm}{\textcolor{red}{E}UMMER}&~~~~~~~\raisebox{.30cm}{SUMMER} \\
\includegraphics[width=2.5cm,height=0.8cm]{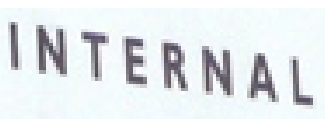} & ~~~~~~~\raisebox{.30cm}{I\textcolor{red}{D}TERNAL} &~~~~~~~\raisebox{.30cm}{I\textcolor{red}{D}TERNAL}&~~~~~~~\raisebox{.30cm}{INTERNAL} \\
\includegraphics[width=2.5cm,height=0.8cm]{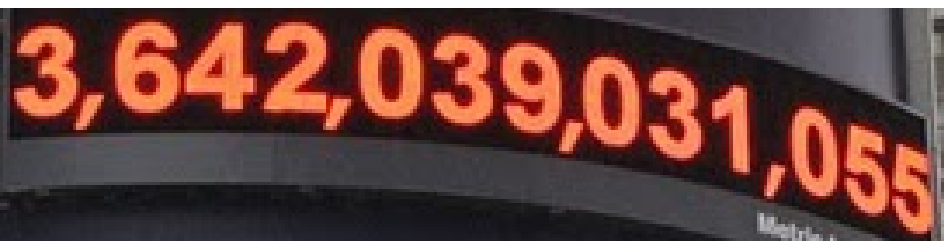} &~~~~~~~\raisebox{.30cm}{364203903105\textcolor{red}{S}}&~~~~~~~\raisebox{.30cm}{3642039031055}&~~~~~~~\raisebox{.30cm}{3642039031055} \\
\includegraphics[width=2.5cm,height=0.8cm]{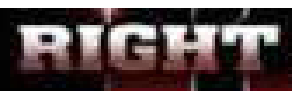} &~~~~~~~\raisebox{.30cm}{R\textcolor{red}{E}GHT}& ~~~~~~~\raisebox{.30cm}{R\textcolor{red}{E}GHT}&~~~~~~~\raisebox{.30cm}{RIGHT} \\
\includegraphics[width=2.5cm,height=0.8cm]{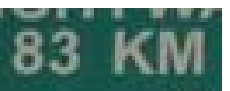} & ~~~~~~~\raisebox{.30cm}{83KM}&~~~~~~~\raisebox{.30cm}{\textcolor{red}{BO}KM}&~~~~~~~\raisebox{.30cm}{\textcolor{red}{BOOM}} \\
\end{tabular}
\caption{Results of our higher order model on a few sample images. Characters
in red represent incorrect recognition. The unary term alone, based on the SVM
classifier, yields poor accuracy, and adding pairwise terms to it improves
this. Due to their limited expressiveness, they do not correct all the errors.
Higher order potentials capture larger context from the English language, and
help address this issue. Note that our method also deals with non-dictionary
words (e.g., second row) and non-horizontal text (sixth row). A typical failure
case containing alphanumeric words is shown in the last row. (\textbf{Best
viewed in colour}).}
\label{fig:word_results}
\end{figure*}

\paragraph{Alternatives for character detection.}
\label{binMethods}
While our sliding window approach for character detection performs well in
several scenarios, including text that is not aligned with the image axes to a
small extent (e.g., rows 4 - 6 in Figure~\ref{fig:word_results}), there are
other alternatives. In particular, we investigated the use of binarization,
MSER~\cite{mser}, and CSER~\cite{NeumannM13} algorithms.
In the first experiment, we replaced our detection module with a binarization
based character extraction scheme -- either a traditional binarization
technique~\cite{otsu} or a more recent random field based
approach~\cite{MishraJ11}. A connected component analysis was performed on the
binarized images to obtain a set of potential character locations. We then
defined the CRF on these characters and performed inference to get the text
contained in the image. These results are summarized in
Table~\ref{tab:binResults}. We observe that binarization based methods perform
poorly compared to our model using a sliding window detector, both in terms of
character-level recall and word recognition. They fail in extracting characters
in the presence of noise, blur or large foreground-background variations. MSER~\cite{mser} or related algorithms (e.g., CSER~\cite{NeumannM13}) may also help to deal with text that is not axis-oriented, but they are not necessarily ideal for character extraction
compared to a sliding window method. To study this, we replaced our sliding window based
character detection scheme with either one of these approaches.
From Table~\ref{tab:binResults} we observe that sliding window character extraction is marginally better than CSER and
significantly better than MSER. One of the reasons for this is that the
classifier used in the sliding window detector is trained on a large variety of
character classes and is less prone to errors than the MSER equivalent. These
results further justify our choice of sliding window based character detection,
although the challenging problem of effectively dealing with text that is not
axis-oriented remains an interesting task for the future.

\paragraph{Effect of pruning}
We propose a pruning step to discard candidates based
on a combination of character-specific aspect ratio and classification scores
(\ref{eq:gs}), instead of simply using extreme aspect ratio to discard character
candidates. This pruning helps in removing many false positive windows, and thus
improves recognition performance. We conducted an experiment to study the
effect of pruning on the IIIT-5K dataset in the open vocabulary setting, and
observed a gain of 4.23\% (46.73\% vs 42.50\%) due to pruning. 

\paragraph{CRF order} 
We varied the order of the CRF from two to six and obtained accuracy of 32\%, 43\%,
45\%, 43\%, 42\% respectively on the IIIT 5K-word dataset in the open vocabulary
setting. Increasing the CRF order beyond four forces a recognized word to be
one from the dictionary, which leads to poor recognition performance for
non-dictionary words, and thus deteriorates the overall accuracy. Empirically,
the fourth order prior shows the best performance.
%May add intuition in the future, anything from English language, for example?

\paragraph{Limits of statistical language models}
Statistical language models have been very useful in improving traditional OCR
performance, but they are indeed limited~\cite{SmithR11,kornai1994}. For
instance, using a large weight for language prior potentials may bias the
recognition towards the closest dictionary word. This is especially true when
the character recognition part of the pipeline is weak. We study such impact of
language models in this experiment. Our analysis on the IIIT 5K-word dataset
suggests that many of the non-dictionary words are composed of valid English
$n$-grams (see Table~\ref{tab:langModel}). However, there are few exceptions,
e.g., words like 35KM, 21P, which are composed of digits and characters; see
last row of Fig.~\ref{fig:word_results}. Using language models has an adverse
effect on the recognition performance in such cases. This results in inferior
recognition performance on non-dictionary words as compared to dictionary words,
e.g. on IIIT-5K dataset our method achieves 51\% and 24\% word recognition accuracy
on dictionary and non-dictionary words respectively.

\section{Summary}
\label{sec:conclusion}
This paper proposes an effective method to recognize scene text. Our model
combines bottom-up cues from character detections and top-down cues from
lexicon. We jointly infer the location of true characters and the word they
represent as a whole. We evaluated our method extensively on several
challenging street scene text datasets, namely SVT, ICDAR 2003/2011/2013, and
IIIT 5K-word and showed that our approach significantly advances the energy
minimization based approach for scene text recognition. In addition to
presenting the word recognition results, we analyzed the different components
of our pipeline, presenting their pros and cons. Finally, we showed that the
energy minimization framework is complementary to the resurgence of
convolutional neural network based techniques, which can help build better
scene understanding systems.
\setlength{\tabcolsep}{2pt}
\begin{table}[!t]
\caption{Character recall (C.\ recall) and recognition accuracy, with unary
only (Unary), unary and pairwise (Pairwise) and the full higher order (H.\
order) models, (all in \%), on the IIIT 5K-word dataset with various character
extraction schemes (Char.\ method). See text for details.}
\centering
\begin{tabular}{|l|c|c|c|c|}
\hline
{Char.\ method} & {C.\ recall} & {Unary} & {Pairwise} & {H.\ order}\\
\hline
Otsu~\cite{otsu}           & 56 & 17.07 & 20.20 & 24.87\\
MRF model~\cite{MishraJ11} & 62 & 20.10 & 22.97 & 28.03\\
MSER~\cite{mser}                    & 72 & 23.20 & 28.50 & 34.70 \\
CSER~\cite{NeumannM13}~\cite{Gomez14}   & 78  & 24.50  & 30.00 & 42.87\\
Sliding window             & {80} & {25.83} & {32.00} & {44.50}\\
\hline
\end{tabular}
\label{tab:binResults}
\end{table}

\paragraph{Acknowledgements}
We thank Jerod Weinman for providing the large lexicon. This work was partially
supported by the Ministry of Communications and Information Technology,
Government of India, New Delhi. Anand Mishra is supported by Microsoft
Corporation and Microsoft Research India under the Microsoft Research India PhD
fellowship award.

%\section*{References}
%\bibliographystyle{elsarticle-num}
\bibliography{mybibfile}

%\begin{table*}[!t]
%%\renewcommand{\arraystretch}{1.3}
%\centering
%\caption{Character classification accuracy (in \%). Layers -- FC1A: 15 FC1B: 14 FC1C: 13 FC1D: 12
%(total layers: 16). $*$We do not evaluate the
%convolutional neural network classifier in~\cite{AZ14} (CNN feat+classifier) on
%the c74K dataset, since the entire dataset was used to train the network.}
%\begin{tabular}{|l|c|c|c|c|}
%\hline
%{Method} & {SVT} & {ICDAR} & {c74K} & {IIIT 5K} \\
%\hline
%H-36+AT+E.F.M+Linear (CI) & 75 & 77 & {79} & 75 \\
%CNN feat+classifier~\cite{AZ14} (CI) & {83} & {86} & $*$ & {85} \\
%CNN feat layer-FC1D+E.F.M+linearSVM (CI) & {81} & {84} & $*$ & {85}  \\
%CNN feat layer-FC1C+E.F.M+linearSVM (CI) & {79} & {84} & $*$ & {84}  \\
%CNN feat layer-FC1B+E.F.M+linearSVM (CI) & {79} & {84} & $*$ & {84}  \\
%CNN feat layer-FC1A+E.F.M+linearSVM (CI) & {78} & {82} & $*$ & {83}  \\
%CNN feat layer-FC1D+linearSVM (CI) & {82} & {84} & $*$ & {85}  \\
%CNN feat layer-FC1C+linearSVM (CI) & {82} & {83} & $*$ & {84}  \\
%CNN feat layer-FC1B+linearSVM (CI) & {81} & {83} & $*$ & {84}  \\
%CNN feat layer-FC1A+linearSVM (CI) & {82} & {84} & $*$ & {85} \\
%\hline
%\end{tabular}
%\end{table*}

\end{document}